%% file: distributed_nonlinear_SA.tex
\documentclass[11pt]{article}
\usepackage{lmodern}

\usepackage[utf8]{inputenc}

\usepackage{geometry}%
\usepackage{graphicx}
\usepackage{amsmath, amsthm, amsfonts, amssymb, graphicx, cite, epsfig,epstopdf,color,soul}

\usepackage{algorithmic,tabularx}
\usepackage[ruled,vlined]{algorithm2e}

\usepackage[bookmarks=true,pageanchor,colorlinks,linkcolor=red,anchorcolor=blue, citecolor=blue,urlcolor=blue,hyperfootnotes=false]{hyperref}

\input{header}
\title{Local Stochastic Approximation: A Unified View of Federated Learning and Distributed Multi-Task Reinforcement Learning Algorithms}
\author{
Thinh T. Doan\thanks{Thinh T. Doan is with the School of Electrical and Computer Engineering, Georgia Institute of Technology, GA, 30332, USA. Email: {\tt\small thinhdoan@gatech.edu}}
}
\date{}

\begin{document}

\maketitle

\begin{abstract}
Motivated by broad applications in reinforcement learning and federated learning, we study local stochastic approximation over a network of agents, where their goal is to find the root of an operator composed of the local operators at the agents. Our focus is to characterize the finite-time performance of this method when the data at each agent are generated from Markov processes, and hence they are dependent. In particularly, we provide the explicit convergence rates of local stochastic approximation for both constant and time-varying step sizes. Our results show that these rates are within a logarithmic factor of the ones under independent data. We then illustrate the applications of these results to different interesting problems in multi-task reinforcement learning and federated learning.  
\end{abstract}

\input{intro}

\input{prob}

\input{analysis}

\input{motivation}

\section{Concluding remark}
This paper studies local stochastic approximation over a network of agents, where the data at each agent are generated from a Markov process. Our main contribution is to provide a finite-time bound for the convergence of mean square errors generated by the algorithm to zero. Our results generalized the existing literature, where the data at the agents are i.i.d, and therefore, the current approach cannot be applied to some algorithms in multi-task reinforcement learning over multi-agent systems.      

\newpage

\bibliographystyle{IEEEtran}
\bibliography{refs}

\input{appendix}

\end{document}

%% file: header.tex
\newcommand{\Eset}{\mathbb{E}}

\newcommand{\Pset}{\mathbb{P}}

\newcommand{\Rset}{\mathbb{R}}

\newcommand{\Acal}{{\cal A}}

\newcommand{\Fcal}{{\cal F}}

\newcommand{\Kcal}{{\cal K}}

\newcommand{\Mcal}{{\cal M}}

\newcommand{\Ocal}{{\cal O}}
\newcommand{\Pcal}{{\cal P}}

\newcommand{\Rcal}{{\cal R}}
\newcommand{\Scal}{{\cal S}}
\newcommand{\Tcal}{{\cal T}}

\newcommand{\Xcal}{{\cal X}}

\newcommand{\Abf}{{\bf A}}

\newcommand{\btheta}{{\bar{\theta}}}

\newtheorem{lem}{Lemma}
\newtheorem{thm}{Theorem}

\newtheorem{assump}{Assumption}
\newtheorem{remark}{Remark}

%% file: intro.tex

\section{Introduction}\label{sec:intro}

In this paper, we study local stochastic approximation ({\sf SA}), a distributed variant of the classic {\sf SA} originally introduced by Robbins and Monro \cite{RobbinsM1951} for solving the root-finding problems under corrupted measurements of an operator (or function). We consider the setting where there are a group of agents communicating indirectly through a centralized coordinator. The goal of the agents is to find the root of the operator, which is composed of the local operators at the agents. For solving this problem, each agent iteratively runs a number of local {\sf SA} steps based on its own data, whose iterates are then averaged at the centralized coordinator. This algorithm, presented in detail in Section \ref{sec:dist_local_SA}, is relatively simple and efficient for solving problems requiring a large amount of data that are distributed to different agents.    

We are motivated by the broad applications of local {\sf SA} in solving problems in federated learning \cite{Kairouz_survery2020,Li_survery2020} and multi-task reinforcement learning \cite{mnih2016asynchronous,espeholt2018impala,hessel2019multi}. These two areas share a common framework, where multiple agents (clients or workers) collaboratively solve a machine learning/reinforcement learning problem under the coordination of a centralized server \cite{Kairouz_survery2020,Li_survery2020,mnih2016asynchronous,espeholt2018impala,hessel2019multi}. In stead of sharing the data collected at the local devices/environments to the server, the agents run local updates of their models/policy based on their data, whose results are then aggregated at the server with the goal to find the global learning objective. In these contexts, local {\sf SA} can be used to formulate the popular algorithms studied in these two areas, as shown in Section \ref{sec:motivating_applications}.   

Our focus, in this paper, is on the theoretical aspects of the finite-time performance of the local stochastic approximation. Our goal is to characterize the convergence rate of this method when the data at the agents are heterogeneous and dependent. In particular, we consider the case when the data at each agent is generated from a Markov process as often considered in the context of multi-task reinforcement learning. Our setting generalizes the existing works in the literature, where the local data is assumed i.i.d. Under fairly standard assumptions, our main contribution is to show that the convergence rates of local {\sf SA} is within a logarithmic factor of the comparable bounds for independent data. As illustrated in Section \ref{sec:motivating_applications}, the  results in this paper provide a unified view for the finite-time bounds of federated learning and multi-task reinforcement learning algorithms under different settings.



\subsection{Related works}
Stochastic approximation is the most efficient and widely used method for solving stochastic optimization problems in many areas, including machine learning \cite{BottouCN2018} and reinforcement learning \cite{SBbook2018,BT_1999_book}. The asymptotic convergence of {\sf SA} under Markov randomness is often done by using the ordinary differential equation ({\sf ODE}) method \cite{borkar2008,KY2009}. Such {\sf ODE} method shows that under the right conditions the noise effects eventually average out and the {\sf SA} iterate asymptotically follows a stable {\sf ODE}. On the other hand, the rates of {\sf SA} have been mostly considered in the context of stochastic gradient descent ({\sf SGD}) under i.i.d samples  \cite{BottouCN2018,Lan_2020_book}. The finite-time convergence of {\sf SGD} under Markov randomness has been studied in \cite{SunSY2018,DoanNPR2020} and the references therein. In the context of reinforcement learning, such results have been studied in \cite{Bhandari2018_FiniteTD, SrikantY2019_FiniteTD, HuS2019,Karimi_colt2019, Doan2019} for linear {\sf SA} and a recent work in \cite{ChenZDMC2019} for nonlinear {\sf SA}.

The local {\sf SA} method considered in this paper has recently received much interests in the context of federated learning  under the name of local {\sf SGD}; see for example \cite{Doan_random_projection_2018,stich2018local,McMahan_LocalSGD17,Woodworth_LocalSGD2020,Haddadpour_LocalSGD2019,Karimireddy_SCAFFOLDSC2019,Khaled_LocalSGD2019}. Finite-time bounds of local {\sf SGD} in these works are derived when the local data at each agent are sampled i.i.d. On the other hand, our focus is to consider the setting where the local data at each agent are sampled from a Markov process, and therefore, they are dependent. We note that the popular distributed {\sf SGD} in machine learning also shares the same communication structure as in the local {\sf SGD}. However, while in local {\sf SGD} the agents have heterogeneous objectives and only share their iterates with the centralized coordinator, in distributed {\sf SGD} the agents compute the stochastic gradients of a global objective and send them to the centralized coordinator.

Finally, distributed/local stochastic approximation has also found broad applications in the context of (multi-task) reinforcement learning 
\cite{espeholt2018impala,hessel2019multi,liu2016decoding, Yu_gradientSurgery2020,mnih2016asynchronous, Arun_massiveRL2015}, which is the main motivation of this paper. In these application, since reinforcement learning is often modeled as a Markov random process, the noise in local {\sf SA} is Markovian. However, there is a lack of understanding about its finite-time performance. Our results in this paper, therefore, help to fill this gap.

%% file: prob.tex

\section{Local stochastic approximation}\label{sec:dist_local_SA}
We consider distributed learning framework, where there are a group of $N$ agents communicating indirectly through a centralized coordinator. Associated with each agent $i$ is a local operator $F_{i}:\Rset^{d}\rightarrow \Rset^{d}$. The goal of the agents is to find the solution $\theta^*$ satisfying
\begin{align}
F(\theta^*) \triangleq \sum_{i=1}^{N}F_{i}(\theta^*) = 0,\label{prob:main_eq}    
\end{align}
where each $F_{i}:\Rset^{d}\rightarrow\Rset^{d}$ is given as 
\begin{align}
F_{i}(\theta) \triangleq \Eset_{\pi_{i}}\left[F_{i}(\theta;X_{i})\right] = \sum_{X_{i}\in\Xcal_{i}}\pi_{i}(X_{i})F_{i}(\theta;X_{i}).\label{prob:Fi}     
\end{align}
Here $\Xcal_{i}$ is a statistical sample space with probability distribution $\pi_{i}$ at agent $i$. We assume that each agent $i$ has access to operator $F_{i}$  only through its samples $\{F_{i}(\cdot,X_{i}^{k})\}$, where $\{X_{i}^{k}\}$ is a sequence of samples of the random variable $X_{i}$. We are interested in the case where each sequence  $\{X_{i}^{k}\}$ is generated from an ergodic Markov process, whose stationary distribution is $\pi_{i}$. Moreover, the sequences $\{X_{i}^{k}\}$ are independent across $i$. For solving problem \eqref{prob:main_eq}, our focus is to study the local stochastic approximation formally stated in Algorithm \ref{alg:Dist_non_SA}. In our algorithm, we implicitly assume that there is an oracle that returns to agent $i$ the value $F_{i}(\theta;X_{i})$ for a given $\theta$ and $X_{i}$.  

\begin{algorithm}[H]
\SetAlgoLined
\textbf{Initialization:} Each agent i initializes $\theta_i^0\in\Rset^d$, a sequence of step sizes $\{\alpha_{k}\}$, and a positive integer $H$.\\ 
The centralized coordinator initializes $\btheta^{0} = 1/N\sum_{i=1}^{N}\theta_i^0$.\\
 \For{k=0,1,2,...}{
  \For {each worker i}{
    1) Receive  $\btheta^{k}$ sent by the centralized coordinator\\
    2) Set $\theta_{i}^{k,0} = \btheta^{k}$\\
    \For {$t = 0,1,\ldots,H-1$}
    {
    \begin{align}
        \theta_{i}^{k,t+1} = \theta_{i}^{k,t} - \alpha_{k+t} F_{i}(\theta_{i}^{k,t};X_{i}^{k+t}).\label{alg:theta_i}
    \end{align}
    }
    }
  The centralized coordinator receives $\theta_{i}^{k,H}$ from each agent $i$ and implements 
  \begin{align}
      \btheta^{k+1} = \frac{1}{N}\sum_{i=1}^{N}\theta_{i}^{k,H}. \label{alg:btheta}
  \end{align}
  }
\caption{Local stochastic approximation}
\label{alg:Dist_non_SA}
\end{algorithm}

Algorithm \ref{alg:Dist_non_SA} is relatively simple to implement, which can be explained as follows. In this algorithm, each agent $i$ maintains a copy $\theta_{i}$ of the solution $\theta^*$ and the centralized coordinator maintains $\btheta$ to estimate the average of $\theta_{i}$. At any iteration $k\geq 0$, each agent $i$ first received $\btheta^{k}$ from the centralized coordinator and initalizes its iterate $\theta_{i}^{k,0} = \btheta^{k}$. Here $\theta_{i}^{k,t}$ denotes the iterate at iteration $k$ and local time $t\in[0,\ldots,H-1]$.  Agent $i$ then runs a number $H$ of local stochastic approximation steps using the time-varying step sizes $\alpha_{k}$ and based on its local data $\{X_{i}^{k+t}\}$. After $H$ local steps, the agents then send their new local updates $\theta_{i}^{k,H}$ to the centralized coordinator to update $\btheta^{k+1}$ by taking the average of these local values.

%% file: analysis.tex

\section{Convergence analysis}
In this section, we study the finite-time performance of Algorithm \ref{alg:Dist_non_SA} when each operator $F_{i}$ is strongly monotone. We provide an upper bound for the convergence rate of the mean square error $\Eset\left[\|\btheta^{k}-\theta^*\|^2\right]$ to zero for both constant and time-varying step sizes. In particular, under constant step size, $\alpha_{k} = \alpha$ for some constant $\alpha$ and $k \gtrsim \log(1/\alpha)$, this convergence occurs at a rate
\begin{align*}
\Eset\left[\|\btheta^{k+1}-\theta^*\|^2\right] \leq  \left(1-\frac{H\mu\alpha}{N}\right)^{k+1-\tau(\alpha)}\Eset\left[\|\btheta^{\tau(\alpha)}-\theta^*\|^2\right] + \Ocal\left(CNHLB^2\log(1/\alpha)\alpha  \right),
\end{align*}
where $\tau(\alpha)$ denotes the mixing time of the underlying Markov chain. On the other hand, under time-varying step sizes $\alpha_{k}\sim 1/(k+1)$ this rate happens at 
\begin{align*}
\Eset\left[\|\btheta^{k+1}-\theta^*\|^2\right] \leq \Ocal\left(\frac{\Eset\left[\|\btheta^{\Kcal^*}-\theta^*\|^2\right]}{k^2}\right) +  \Ocal\left(\frac{CHLB^2\log(k)}{k}\right), \qquad \forall k\geq \Kcal^*   
\end{align*}
for some positive integer $\Kcal^*$ depending on the problem parameters $C,L,B,\mu$, which we will define shortly. First, our rates scale linearly with the number  of local steps $H$. As expected, when $H$ goes to $\infty$, each agent only consider its local {\sf SA} without communicating with the centralized  coordinator. In this case, one would expect that each agent only finds the root of its own operator. Second, one can view the constant $B$ as the variance of the noise. Finally, our rates match the ones of local {\sf SGD} under i.i.d noise \cite{Khaled_LocalSGD2019}, except for a $\log$ factor reflecting the Markovian randomness. 

We start our analysis by introducing the following technical assumptions and notation used in this section. Given a constant $\alpha > 0$, we denote by $\tau_{i}(\alpha)$ the mixing time associated with the Markov chain $\{X_{i}^{k}\}$, i.e., the following condition holds  
\begin{align}
\|\Pset_{i}^{k}(X_{i}^{0},\cdot) - \pi_{i} \|_{TV} \leq \alpha,\quad \forall k\geq \tau_{i}(\alpha),\;\forall X_{i}^{0}\in\Xcal_{i},\;\forall i\in[N],\label{notation:tau}        
\end{align}
where  $\|\cdot\|_{TV}$ is the total variance distance and $\Pset_{i}^{k}(X_{i}^{0},X_{i})$ is the probability that $X_{i}^{k} = X_{i}$ when we start from $X_{i}^{0}$. The mixing time represents the time $X_{i}^{k}$ getting close to the stationary distribution $\pi_{i}$. In addition, we denote by $\tau(\alpha) = \max_{i}\tau_{i}(\alpha)$. 

For our results derived in this section, we consider the following assumptions. For simplicity, we assume that these assumptions hold to the rest of this paper. 
\begin{assump}\label{assump:ergodicity}
The Markov chain $\{X_{i}^{k}\}$ is erogodic (irreducible and aperiodic) with finite state space $\Xcal_{i}$. 
\end{assump}
\begin{assump}\label{assump:Lipschitz_sample}
The mapping $F_{i}(\cdot)$ and $F_{i}(\cdot,X_{i})$ is Lipschitz continuous in $\theta$ almost surely, i.e., there exists a positive constant $L$ such that for all $X_{i}\in\Xcal_{i}$ and $i\in[N]$
\begin{align}
\|F_{i}(\theta)-F_{i}(\omega)\|\leq L\|\theta-\omega\|\quad \text{and}\quad \|F_{i}(\theta,X_{i}) - F_{i}(\omega,X_{i})\|   \leq L\|\theta-\omega\|,\qquad  \forall \theta,\omega\in\Rset^{d}.  \label{assump:Lipschitz_sample:ineq}
\end{align}
\end{assump}
\begin{assump}\label{assump:strong_monotone}
There exists a positive constant $\mu$ such that
\begin{align}
(F_{i}(\theta) - F_{i}(\omega))^T(\theta - \omega) \geq \mu\|\theta-\omega\|^2,\qquad  \forall\theta,\omega\in\Rset^{d},\;\forall i\in[N].   \label{assump:strong_monotone:ineq}  
\end{align}
\end{assump}
Assumption \ref{assump:ergodicity} implies that the Markov chain $\{X_{i}^{k}\}$ has geometric mixing time, which depends on the second largest eigenvalue of the transition probability matrix of $\{X_{i}^{k}\}$ \cite{LevinPeresWilmer2006}. This assumption holds in various applications, e.g, in incremental optimization \cite{RamNV2009} and in reinforcement learning modeled by Markov decision processes with a finite number of states and actions \cite{silver2017mastering}. In addition, Assumption  \ref{assump:ergodicity} is used in the existing literature to study the finite-time performance of {\sf SA} and its distributed variants under Markov randomness; see  \cite{SunSY2018,  SrikantY2019_FiniteTD, ChenZDMC2019, Doan2019, wu2020finite, DoanMR2019_DTD} and the references therein.

Assumption \ref{assump:Lipschitz_sample} is often used in local {\sf SGD} methods  in federated learning \cite{Khaled_LocalSGD2019}. This assumption also holds in the context of reinforcement learning considered in Section \ref{sec:motivating_applications}. Assumption \ref{assump:strong_monotone} implies that $F_{i}$ are strongly monotone (or strongly convex in the context of {\sf SGD}).

The following result is a consequence of Assumptions \ref{assump:ergodicity} and \ref{assump:Lipschitz_sample}, which is shown in \cite[Lemma $3.2$]{ChenZDMC2019}. This lemma basically states that each Markov chain $X_{i}$ has a geometric mixing time, which translates to the operator $F_{i}$ due to the Lipschitz condition.   
\begin{lem}\label{lem:mixing_time}
There exists a constant $C>0$ such that given $\alpha > 0$ we have, $\forall$ $i\in[N]$, $\tau_{i}\leq C\log(1/\alpha)$ and 
\begin{align}
\left|\Eset\left[F_{i}(\theta,X_{i}^{k})\,|\,X_{i}^{0}\right] - F_{i}(\theta)\right| \leq \alpha(\|\theta\|+1),\qquad \forall \theta,\;\forall k\geq \tau_{i}(\alpha).    \label{lem:mixing_time:ineq}
\end{align}
\end{lem}
Finally, since $\Xcal_{i}$ is finite for all $i\in[N]$, Assumption \ref{assump:Lipschitz_sample} also gives the following result. 
\begin{lem}\label{lem:bounded_Fi}    
Let $B = \max_{i}\max_{X_{i}\in\Xcal_{i}}\|F_{i}(0,X_{i})\|$. Then we have for all $\theta\in 
\Rset^{d}$
\begin{align}
\|F_{i}(\theta)\| \leq B(\|\theta\|+1)\quad \text{and}\quad  \|F_{i}(\theta,X_{i})\| \leq B(\|\theta\| + 1),\qquad \forall X_{i}\in\Xcal_{i},\;\forall i\in[N].\label{lem:bounded_Fi:ineq}   
\end{align}
\end{lem}

\subsection{Constant step sizes}\label{sec:analysis:const}
In this subsection, we derive the rate of Algorithm \ref{alg:Dist_non_SA} under constant step sizes, that is, $\alpha_{k} = \alpha$ for all $k\geq 0$. Note that by Lemma \ref{lem:mixing_time} $\tau_{i}(\alpha) \leq C\log(1/\alpha)$ given a constant $\alpha > 0$. Thus, we have  $\lim_{\alpha\rightarrow0}\tau_{i}(\alpha)\alpha = 0$ for all $i\in[N]$. This implies that there exists a sufficiently small positive $\alpha$ such that
\begin{align}
\alpha\tau(\alpha)\leq \min\left\{\frac{\log(2)}{2BH}\;,\;\frac{\mu}{8N\left(19B^2H + 9 + 57LBH \right)}\;,\;\frac{N}{2H\mu}\right\}, \label{sec:analysis:const:stepsize}
\end{align} 
where recall that $\tau(\alpha) = \max_{i}\tau_{i}(\alpha)$. The result in this section is established under condition \eqref{sec:analysis:const:stepsize}.  Under constant step sizes, we have from \eqref{alg:theta_i} for all $k\geq 0$
\begin{align}
\theta_{i}^{k,H} = \theta_{i}^{k,0} - \alpha\sum_{t=0}^{H-1}F_{i}(\theta_{i}^{k,t};X_{i}^{k+t}) = \btheta^{k} - \alpha\sum_{t=0}^{H-1}F_{i}(\theta_{i}^{k,t};X_{i}^{k+t}),\label{sec:analysis:const:theta_iH}      
\end{align}
which implies
\begin{align}
\btheta^{k+1} = \btheta^{k} - \frac{\alpha}{N}\sum_{i=1}^{N}\sum_{t=0}^{H-1}F_{i}(\theta_{i}^{k,t};X_{i}^{k+t}).\label{sec:analysis:const:btheta_k}    
\end{align}
To derive the finite-time bound of Algorithm \ref{alg:Dist_non_SA}, we require the following three technical lemmas. For an ease of exposition, their proofs are presented in the Appendix. The first lemma is to upper bound the norm of $\theta_{i}$ by the norm of $\btheta$. 
\begin{lem}\label{sec:analysis:const:lem:theta_i_bound}
Let $\{\btheta^{k}\}$ and $\{\theta_{i}^{k,t}\}$, for all k $\geq0$ and $t\in[1,H]$, be generated by Algorithm \ref{alg:Dist_non_SA}. In addition, let the step size $\alpha$ satisfy \eqref{sec:analysis:const:stepsize}. Then the following relations hold for all $k\geq 0$ and $t\in[0,H]$  
\begin{align}
\|\theta_{i}^{k,t}\| &\leq 2\|\btheta^{k}\| + 2BH\alpha \leq 2\|\btheta^{k}\| + 1. \label{sec:analysis:const:lem:theta_i_bound:ineq1}\\
\|\theta_{i}^{k,t} - \btheta^{k}\| &\leq 2BH\alpha\|\btheta^{k}\| + 2BH\alpha.\label{sec:analysis:const:lem:theta_i_bound:ineq2}
\end{align}
\end{lem}
Our next lemma is to provide an upper bound for the quantity $\|\btheta^{k}-\btheta^{k-\tau(\alpha)}\|$.
\begin{lem}\label{sec:analysis:const:lem:btheta_k_tau}
Let all the conditions in Lemma \ref{sec:analysis:const:lem:theta_i_bound} hold. Then the following relations hold $\forall k\geq 0$ and $t\in[0,H]$  
\begin{align}
\|\btheta^{k}-\btheta^{k-\tau(\alpha)}\| &\leq  12BH\alpha\tau(\alpha)\|\btheta^{k}\| + 12BH\alpha\tau(\alpha)\leq 2\|\btheta^{k}\| + 2. \label{sec:analysis:const:lem:btheta_k_tau:ineq1} \\
\|\btheta^{k}-\btheta^{k-\tau(\alpha)}\|^2 &\leq    288B^2H^2\alpha^2\tau^2(\alpha)\|\btheta^{k}\|^2 + 288B^2H^2\alpha^2\tau^2(\alpha)\leq 8\|\btheta^{k}\|^2 + 8\label{sec:analysis:const:lem:btheta_k_tau:ineq2}. 
\end{align}
\end{lem}
Finally, we present an upper  bound for the bias caused by Markovian noise. 
\begin{lem}\label{sec:analysis:const:lem:bias}
Let all the conditions in Lemma \ref{sec:analysis:const:lem:theta_i_bound} hold. Then we have
\begin{align}
-\sum_{i=1}^{N}\sum_{t=0}^{H-1}\Eset\left[\left\langle\btheta^{k}-\theta^*,F_{i}(\theta_{i}^{k,t};X_{i}^{k+t})-F_{i}(\btheta^{k})\right\rangle\right] &\leq 36NH\alpha\Eset\left[\|\btheta^{k}-\theta^*\|^2\right] + 36NH\alpha \left(1 + \|\theta^*\|\right)^2\notag\\ 
&\qquad + 12(19L+6B)NBH^2\alpha\tau(\alpha)\Eset\left[\|\btheta^{k}-\theta^*\|^2\right]\notag\\ 
&\qquad + 12(19L+6B)NBH^2\left(1 + \|\theta^*\|\right)^2\alpha\tau(\alpha) \label{sec:analysis:const:lem:bias:ineq}.
\end{align}
\end{lem} 
Our first main result in this paper is presented in the following Theorem, where we derive the rate of Algorithm \ref{alg:Dist_non_SA} under constant step sizes. 
\begin{thm}\label{thm:const}
Let $\{\btheta^{k}\}$ and $\{\theta_{i}^{k,t}\}$, for all k $\geq0$ and $t\in[1,H]$, be generated by Algorithm \ref{alg:Dist_non_SA}. In addition, let the step size $\alpha$ satisfy \eqref{sec:analysis:const:stepsize}. Then we have
\begin{align}
\Eset\left[\|\btheta^{k+1}-\theta^*\|^2\right]
&\leq \left(1-\frac{H\mu\alpha}{N}\right)^{k+1-\tau(\alpha)}\Eset\left[\|\btheta^{\tau(\alpha)}-\theta^*\|^2\right]\notag\\
&\qquad + \frac{8NC\left(B^2H + 9 + 3(19L+6B)BH \right)(\|\theta^*\|+1)^2}{\mu}\log(1/\alpha)\alpha.    \label{thm:const:ineq}
\end{align}
\end{thm}
\begin{remark}
Under constant step sizes, the rate in \eqref{thm:const:ineq} shows that the mean square errors generated by Algorithm \ref{alg:Dist_non_SA} decays to a ball surrounding the origin exponentially. As $\alpha$ decays to zero, this error also goes to zero. Second, our rate is only different from the one using i.i.d data by a $\log$ factor, for example in \cite{Khaled_LocalSGD2019}. This reflects the impact of Markovian randomness through the mixing time $\tau(\alpha)$. Third, our upper bound scales linearly on the number of local steps $H$.  
\end{remark}
\begin{proof}
By \eqref{sec:analysis:const:btheta_k} we consider
\begin{align}
\|\btheta^{k+1} - \theta^*\|^2 &= \left\|\btheta^{k}-\theta^* - \frac{\alpha}{N}\sum_{i=1}^{N}\sum_{t=0}^{H-1}F_{i}(\theta_{i}^{k,t};X_{i}^{k+t})\right\|^2\notag\\
&= \|\btheta^{k}-\theta^*\|^2 + \left\|\frac{\alpha}{N}\sum_{i=1}^{N}\sum_{t=0}^{H-1}F_{i}(\theta_{i}^{k,t};X_{i}^{k+t})\right\|^2 - \frac{2\alpha}{N}\left\langle\btheta^{k}-\theta^*,\sum_{i=1}^{N}\sum_{t=0}^{H-1}F_{i}(\theta_{i}^{k,t};X_{i}^{k+t})\right\rangle\notag\\
&= \|\btheta^{k}-\theta^*\|^2 + \left\|\frac{\alpha}{N}\sum_{i=1}^{N}\sum_{t=0}^{H-1}F_{i}(\theta_{i}^{k,t};X_{i}^{k+t})\right\|^2 - \frac{2H\alpha}{N}\left\langle\btheta^{k}-\theta^*,F(\btheta^{k})\right\rangle\notag\\
&\qquad - \frac{2\alpha}{N}\left\langle\btheta^{k}-\theta^*,\sum_{i=1}^{N}\sum_{t=0}^{H-1}F_{i}(\theta_{i}^{k,t};X_{i}^{k+t})-F_{i}(\btheta^{k})\right\rangle.\label{lem:opt_dist:Eq1}
\end{align}
First, using \eqref{lem:bounded_Fi:ineq} and \eqref{sec:analysis:const:lem:theta_i_bound:ineq1} we consider the second term on the right-hand side of \eqref{lem:opt_dist:Eq1}
\begin{align}
&\left\|\frac{\alpha}{N}\sum_{i=1}^{N}\sum_{t=0}^{H-1}F_{i}(\theta_{i}^{k,t};X_{i}^{k+t})\right\|^2 \leq \frac{\alpha^2}{N^2}\left|\sum_{i=1}^{N}\sum_{t=0}^{H-1}B(\|\theta_{i}^{k,t}\|+1)\right|^2\leq \frac{\alpha^2}{N^2}\left|\sum_{i=1}^{N}\sum_{t=0}^{H-1}2B(\|\btheta^{k}\|+1)\right|^2\notag\\
&\qquad\leq 4B^2H^2\alpha^2(\|\btheta^{k}\|+1)^2 \leq 8B^2H^2\alpha^2\|\btheta^{k}-\theta^*\|^2 + 8B^2H^2(\|\theta^*\|+1)^2\alpha^2. \label{lem:opt_dist:Eq1a}
\end{align}
Second, by Assumption \ref{assump:strong_monotone} we have
\begin{align}
 - \frac{2H\alpha}{N}\left\langle\btheta^{k}-\theta^*,F(\btheta^{k})\right\rangle \leq -\frac{2H\mu \alpha}{N}\|\btheta^{k}-\theta^*\|^2.\label{lem:opt_dist:Eq1b}    
\end{align}
Thus, taking the expectation on both sides of \eqref{lem:opt_dist:Eq1} and using \eqref{sec:analysis:const:lem:bias:ineq}, \eqref{lem:opt_dist:Eq1a}, and \eqref{lem:opt_dist:Eq1b} yields 
\begin{align*}
\Eset\left[\|\btheta^{k+1}-\theta^*\|\right]&\leq \left(1-\frac{2H\mu\alpha}{N}\right)\Eset\left[\|\btheta^{k}-\theta^*\|^2\right] + 8B^2H^2\alpha^2\Eset\left[\|\btheta^{k}-\theta^*\|^2\right] + 8B^2H^2(\|\theta^*+1\|)^2\alpha^2\notag\\
&\qquad + 72H\alpha^2\Eset\left[\|\btheta^{k}-\theta^*\|^2\right] + 72H\alpha^2 \left(1 + \|\theta^*\|\right)^2\notag\\ 
&\qquad + 24(19L+6B)BH^2\alpha^2\tau(\alpha)\Eset\left[\|\btheta^{k}-\theta^*\|^2\right]\notag\\ 
&\qquad + 24(19L+6B)BH^2\left(1 + \|\theta^*\|\right)^2\alpha^2\tau(\alpha)\notag\\
&\leq \left(1-\frac{2H\mu\alpha}{N}\right)\Eset\left[\|\btheta^{k}-\theta^*\|^2\right]\notag\\ 
&\qquad + 8H\left(B^2H + 9 + 3(19L+6B)BH \right)\tau(\alpha)\alpha^2\Eset\left[\|\btheta^{k}-\theta^*\|^2\right]\notag\\
&\qquad + 8H\left(B^2H + 9 + 3(19L+6B)BH \right)(\|\theta^*\|+1)^2\tau(\alpha)\alpha^2. 
\end{align*}
Recall that $\alpha$ satisfies \eqref{sec:analysis:const:stepsize} and by Lemma \ref{lem:mixing_time} $\tau(\alpha) \leq C\log(1/\alpha)$. Then the preceding relation yields \eqref{thm:const:ineq}, i.e., for all $k\geq \tau(\alpha)$
\begin{align*}
\Eset\left[\|\btheta^{k+1}-\theta^*\|\right]
&\leq \left(1-\frac{H\mu\alpha}{N}\right)\Eset\left[\|\btheta^{k}-\theta^*\|^2\right]\notag\\ 
&\qquad + 8H\left(B^2H + 9 + 3(19L+6B)BH \right)(\|\theta^*\|+1)^2\tau(\alpha)\alpha^2\notag\\ 
&\leq \left(1-\frac{H\mu\alpha}{N}\right)^{k+1-\tau(\alpha)}\Eset\left[\|\btheta^{\tau(\alpha)}-\theta^*\|^2\right]\notag\\ &\qquad + 8H\left(B^2H + 9 + 3(19L+6B)BH \right)(\|\theta^*\|+1)^2\tau(\alpha)\alpha^2\sum_{t=\tau(\alpha)}^{k}\left(1-\frac{H\mu\alpha}{N}\right)^{k-t}\notag\\
&\leq \left(1-\frac{H\mu\alpha}{N}\right)^{k+1-\tau(\alpha)}\Eset\left[\|\btheta^{\tau(\alpha)}-\theta^*\|^2\right]\notag\\
&\qquad + \frac{8NC\left(B^2H + 9 + 3(19L+6B)BH \right)(\|\theta^*\|+1)^2}{\mu}C\log(1/\alpha)\alpha.
\end{align*}
\end{proof}

\subsection{Time-varying step sizes}
In this section, we derive the finite-time bound of Algorithm \ref{alg:Dist_non_SA} under time-varying step sizes $\alpha_{k}$. We consider $\alpha_{k}$ being nonnegative, decreasing, and $\lim_{k\rightarrow \infty}\alpha_{k} = 0$. Thus, by Lemma \ref{lem:mixing_time} we have  $\lim_{k\rightarrow\infty}\tau_{i}(\alpha_{k})\alpha_{k} = 0$ for all $i\in[N]$. This implies that there exists a postive integer $\Kcal^*$ such that for all $k\geq \Kcal^*$
\begin{align}
\sum_{t=k-\tau(\alpha_{k})}^{k}\alpha_{t}\leq \alpha_{k-\tau(\alpha_{k})}\tau(\alpha_{k})\leq \min\left\{\frac{\log(2)}{2BH}\;,\;\frac{\mu}{8N\left(19B^2H + 9 + 57LBH \right)}\;,\;2\alpha_{0}\right\}, \label{sec:analysis:tv:stepsize}
\end{align} 
where recall that $\tau(\alpha_{k}) = \max_{i}\tau_{i}(\alpha_{k})$. For convenience, we denote by 
\begin{align}
\alpha_{k;\tau(\alpha_{k})}    = \sum_{t=k-\tau(\alpha_{k})}^{k}\alpha_{t}. \label{notation:alpha_k_tau_k}
\end{align}
Under $\alpha_{k}$, we have from \eqref{alg:theta_i} for all $k\geq 0$
\begin{align}
\theta_{i}^{k,H} = \theta_{i}^{k,0} - \sum_{t=0}^{H-1}\alpha_{k+t}F_{i}(\theta_{i}^{k,t};X_{i}^{k+t}) = \btheta^{k} -  \sum_{t=0}^{H-1}\alpha_{k+t}F_{i}(\theta_{i}^{k,t};X_{i}^{k+t}),\label{sec:analysis:tv:theta_iH}      
\end{align}
which implies
\begin{align}
\btheta^{k+1} = \btheta^{k} - \frac{1}{N}\sum_{i=1}^{N}\sum_{t=0}^{H-1}\alpha_{k+t}F_{i}(\theta_{i}^{k,t};X_{i}^{k+t}).\label{sec:analysis:tv:btheta_k}    
\end{align}
Similar to the case of constant step sizes, we first consider the following three technical lemmas that are useful for our main result presented in Theorem \ref{thm:tv} below. For an ease of exposition, their proofs are presented in the Appendix. The first lemma is to upper bound the norm of $\theta_{i}$ by the norm of $\btheta$. 
\begin{lem}\label{sec:analysis:tv:lem:theta_i_bound}
Let $\{\btheta^{k}\}$ and $\{\theta_{i}^{k,t}\}$, for all k $\geq0$ and $t\in[1,H]$, be generated by Algorithm \ref{alg:Dist_non_SA}. In addition, let the step size $\alpha$ satisfy \eqref{sec:analysis:tv:stepsize}. Then the following relations hold for all $k\geq 0$ and $t\in[0,H]$  
\begin{align}
\|\theta_{i}^{k,t}\| &\leq 2\|\btheta^{k}\| + 2BH\alpha_{k} \leq 2\|\btheta^{k}\| + 1. \label{sec:analysis:tv:lem:theta_i_bound:ineq1}\\
\|\theta_{i}^{k,t} - \btheta^{k}\| &\leq 2BH\alpha_{k}\|\btheta^{k}\| + 2BH\alpha_{k}.\label{sec:analysis:tv:lem:theta_i_bound:ineq2}
\end{align}
\end{lem}
Our next lemma is to provide an upper bound for the quantity $\|\btheta^{k}-\btheta^{k-\tau(\alpha)}\|$.
\begin{lem}\label{sec:analysis:tv:lem:btheta_k_tau}
Let all the conditions in Lemma \ref{sec:analysis:tv:lem:theta_i_bound} hold. Then the following relations hold $\forall k\geq 0$ and $t\in[0,H]$  
\begin{align}
\|\btheta^{k}-\btheta^{k-\tau(\alpha)}\| &\leq  12BH\alpha_{k;\tau(\alpha_{k})}\|\btheta^{k}\| + 12BH\alpha_{k;\tau(\alpha_{k})}\leq 2\|\btheta^{k}\| + 2. \label{sec:analysis:tv:lem:btheta_k_tau:ineq1} \\
\|\btheta^{k}-\btheta^{k-\tau(\alpha)}\|^2 &\leq    288B^2H^2\alpha_{k;\tau(\alpha_{k})}^2\|\btheta^{k}\|^2 + 288B^2H^2\alpha_{k;\tau(\alpha_{k})}^2\leq 8\|\btheta^{k}\|^2 + 8\label{sec:analysis:tv:lem:btheta_k_tau:ineq2}. 
\end{align}
\end{lem}
Finally, we present an upper  bound for the bias caused by Markovian noise. 
\begin{lem}\label{sec:analysis:tv:lem:bias}
Let all the conditions in Lemma \ref{sec:analysis:tv:lem:theta_i_bound} hold. Then we have
\begin{align}
-\sum_{i=1}^{N}\sum_{t=0}^{H-1}\alpha_{k+t}\Eset\left[\left\langle\btheta^{k}-\theta^*,F_{i}(\theta_{i}^{k,t};X_{i}^{k+t})-F_{i}(\btheta^{k})\right\rangle\right]& \leq 36NH\alpha_{k}^2\Eset\left[\|\btheta^{k}-\theta^*\|^2\right] + 36NH\alpha_{k}^2 \left(1 + \|\theta^*\|\right)^2\notag\\ 
&\quad + 12(19L+6B)NBH^2\alpha_{k}\alpha_{k;\tau(\alpha_{k})}\Eset\left[\|\btheta^{k}-\theta^*\|^2\right]\notag\\ 
&\quad + 12(19L+6B)NBH^2\left(1 + \|\theta^*\|\right)^2\alpha_{k}\alpha_{k;\tau(\alpha_{k})} \label{sec:analysis:tv:lem:bias:ineq}.
\end{align}
\end{lem} 
The second main result in this paper is presented in the following theorem, where we study the rate of Algorithm \ref{alg:Dist_non_SA} under time-varying step sizes. 
\begin{thm}\label{thm:tv}
Let $\{\btheta^{k}\}$ and $\{\theta_{i}^{k,t}\}$, for all k $\geq0$ and $t\in[1,H]$, be generated by Algorithm \ref{alg:Dist_non_SA}. In addition, let the step size $\alpha_{k} = \alpha/(k+1)$ satisfy \eqref{sec:analysis:tv:stepsize} where $\alpha = 2N/(H\mu)$ . Then we have for all $k\geq \Kcal^*$
\begin{align}
\Eset\left[\|\btheta^{k+1}-\theta^*\|^2\right]
&\leq \frac{(\Kcal^*)^2\Eset\left[\|\btheta^{\Kcal^*}-\theta^*\|^2\right]}{(k+1)^2}\notag\\ 
&\qquad + \frac{16\alpha^2 C H\left(B^2H + 9 + 3(19L+6B)BH \right)(\|\theta^*\|+1)^2\log\left(\frac{k+1}{\alpha}\right)}{k+1}\cdot \label{thm:tv:ineq}
\end{align}
\end{thm}
\begin{remark}
Here, we have the same observation as the one in Theorem \ref{thm:const}, except now the rate is sublinear due to time-varying step sizes. However, the mean square errors decay to zero instead of to a neighborhood of the origin.    
\end{remark}
\begin{proof}
By \eqref{sec:analysis:const:btheta_k} we consider
\begin{align}
\|\btheta^{k+1} - \theta^*\|^2 &= \left\|\btheta^{k}-\theta^* - \frac{1}{N}\sum_{i=1}^{N}\sum_{t=0}^{H-1}\alpha_{k+t}F_{i}(\theta_{i}^{k,t};X_{i}^{k+t})\right\|^2\notag\\
&= \|\btheta^{k}-\theta^*\|^2 + \left\|\frac{1}{N}\sum_{i=1}^{N}\sum_{t=0}^{H-1}\alpha_{k+t}F_{i}(\theta_{i}^{k,t};X_{i}^{k+t})\right\|^2 - \frac{2}{N}\left\langle\btheta^{k}-\theta^*,\sum_{i=1}^{N}\sum_{t=0}^{H-1}\alpha_{k+t}F_{i}(\theta_{i}^{k,t};X_{i}^{k+t})\right\rangle\notag\\
&= \|\btheta^{k}-\theta^*\|^2 + \left\|\frac{1}{N}\sum_{i=1}^{N}\sum_{t=0}^{H-1}\alpha_{k+t}F_{i}(\theta_{i}^{k,t};X_{i}^{k+t})\right\|^2 - \frac{2H}{N}\left\langle\btheta^{k}-\theta^*,F(\btheta^{k})\sum_{t=0}^{H-1}\alpha_{k+t}\right\rangle\notag\\
&\qquad - \frac{2}{N}\left\langle\btheta^{k}-\theta^*,\sum_{i=1}^{N}\sum_{t=0}^{H-1}\alpha_{k+t}\left(F_{i}(\theta_{i}^{k,t};X_{i}^{k+t})-F_{i}(\btheta^{k})\right)\right\rangle.\label{thm:tv:Eq1}
\end{align}
First, using \eqref{lem:bounded_Fi:ineq} and \eqref{sec:analysis:tv:lem:theta_i_bound:ineq1} we consider the second term on the right-hand side of \eqref{thm:tv:Eq1}
\begin{align}
&\left\|\frac{1}{N}\sum_{i=1}^{N}\sum_{t=0}^{H-1}\alpha_{k+t}F_{i}(\theta_{i}^{k,t};X_{i}^{k+t})\right\|^2 \leq \frac{\alpha_{k}^2}{N^2}\left|\sum_{i=1}^{N}\sum_{t=0}^{H-1}B(\|\theta_{i}^{k,t}\|+1)\right|^2\leq \frac{\alpha_{k}^2}{N^2}\left|\sum_{i=1}^{N}\sum_{t=0}^{H-1}2B(\|\btheta^{k}\|+1)\right|^2\notag\\
&\qquad\leq 4B^2H^2\alpha_{k}^2(\|\btheta^{k}\|+1)^2 \leq 8B^2H^2\alpha_{k}^2\|\btheta^{k}-\theta^*\|^2 + 8B^2H^2(\|\theta^*+1\|)^2\alpha_{k}^2. \label{thm:tv:Eq1a}
\end{align}
Second, by Assumption \ref{assump:strong_monotone} we have
\begin{align}
 - \frac{2H}{N}\left\langle\btheta^{k}-\theta^*,F(\btheta^{k})\sum_{t=0}^{H-1}\alpha_{k+t}\right\rangle \leq -\frac{2H\mu \sum_{t=0}^{H-1}\alpha_{k+t}}{N}\|\btheta^{k}-\theta^*\|^2\leq \frac{-2H\mu}{N}\alpha_{k}\|\btheta^{k}-\theta^*\|^2.\label{thm:tv:Eq1b}    
\end{align}
Thus, taking the expectation on both sides of \eqref{thm:tv:Eq1} and using \eqref{sec:analysis:tv:lem:bias:ineq}, \eqref{thm:tv:Eq1a}, and \eqref{thm:tv:Eq1b} yields 
\begin{align}
\Eset\left[\|\btheta^{k+1}-\theta^*\|\right]&\leq \left(1-\frac{2H\mu\alpha_{k}}{N}\right)\Eset\left[\|\btheta^{k}-\theta^*\|^2\right] + 8B^2H^2\alpha_{k}^2\Eset\left[\|\btheta^{k}-\theta^*\|^2\right] + 8B^2H^2(\|\theta^*+1\|)^2\alpha_{k}^2\notag\\
&\qquad + 72H\alpha_{k}^2\Eset\left[\|\btheta^{k}-\theta^*\|^2\right] + 72H\alpha_{k}^2 \left(1 + \|\theta^*\|\right)^2\notag\\ 
&\qquad + 24(19L+6B)BH^2\alpha_{k}\alpha_{k;\tau(\alpha_{k})}\Eset\left[\|\btheta^{k}-\theta^*\|^2\right]\notag\\ 
&\qquad + 24(19L+6B)BH^2\left(1 + \|\theta^*\|\right)^2\alpha_{k}\alpha_{k;\tau(\alpha_{k})}\notag\\
&\leq \left(1-\frac{2H\mu\alpha_{k}}{N}\right)\Eset\left[\|\btheta^{k}-\theta^*\|^2\right]\notag\\ 
&\qquad + 8H\left(B^2H + 9 + 3(19L+6B)BH \right)\alpha_{k}\alpha_{k;\tau(\alpha_{k})}\Eset\left[\|\btheta^{k}-\theta^*\|^2\right]\notag\\
&\qquad + 8H\left(B^2H + 9 + 3(19L+6B)BH \right)(\|\theta^*\|+1)^2\alpha_{k}\alpha_{k;\tau(\alpha_{k})}. \label{thm:tv:Eq1c}  
\end{align}
Recall that $\alpha_{k} = \alpha/(k+1)$ satisfies \eqref{sec:analysis:tv:stepsize} where $\alpha = 2N/(H\mu)$. Then we have
\begin{align*}
(k+1)^2\left(1-\frac{H\mu\alpha_{k}}{N}\right)= (k+1)(k-1)\leq k^2.     
\end{align*}
In addition, by Lemma \ref{lem:mixing_time} and \eqref{sec:analysis:tv:stepsize} we have 
\begin{align*}
(k+1)\alpha_{k;\tau(\alpha_{k})} \leq \frac{\alpha\tau(\alpha_{k})(k+1)}{k+1-\tau(\alpha_{k})}\leq 2\alpha\tau(\alpha_{k})\leq 2C\alpha\log\left(\frac{k+1}{\alpha}\right)    
\end{align*}
Thus, multiply both sides of \eqref{thm:tv:Eq1c} by $(k+1)^2$ yields
\begin{align*}
(k+1)^2\Eset\left[\|\btheta^{k+1}-\theta^*\|\right]
&\leq (k+1)^2\left(1-\frac{H\mu\alpha_{k}}{N}\right)\Eset\left[\|\btheta^{k}-\theta^*\|^2\right]\notag\\ 
&\qquad + 8H\left(B^2H + 9 + 3(19L+6B)BH \right)(\|\theta^*\|+1)^2\alpha_{k}\alpha_{k;\tau(\alpha_{k})}(k+1)^2\notag\\ 
&\leq k^2\Eset\left[\|\btheta^{k}-\theta^*\|^2\right]\notag\\ 
&\qquad + 8\alpha H\left(B^2H + 9 + 3(19L+6B)BH \right)(\|\theta^*\|+1)^2\alpha_{k;\tau(\alpha_{k})}(k+1)\notag\\
&\leq k^2\Eset\left[\|\btheta^{\Kcal^*}-\theta^*\|^2\right]\notag\\ 
&\qquad + 16\alpha^2 C H\left(B^2H + 9 + 3(19L+6B)BH \right)(\|\theta^*\|+1)^2\log\left(\frac{k+1}{\alpha}\right)\notag\\
&\leq (\Kcal^*)^2\Eset\left[\|\btheta^{\Kcal^*}-\theta^*\|^2\right]\notag\\ 
&\qquad +  16\alpha^2 C H\left(B^2H + 9 + 3(19L+6B)BH \right)(\|\theta^*\|+1)^2\sum_{t=\Kcal^*}^{k}\log\left(\frac{t+1}{\alpha}\right)\notag\\
&\leq (\Kcal^*)^2\Eset\left[\|\btheta^{\Kcal^*}-\theta^*\|^2\right]\notag\\ 
&\qquad + 16\alpha^2 C H\left(B^2H + 9 + 3(19L+6B)BH \right)(\|\theta^*\|+1)^2k\log\left(\frac{k+1}{\alpha}\right),
\end{align*}
which dividing both sides by $(k+1)^2$ yields \eqref{thm:tv:ineq}.
\end{proof}

%% file: motivation.tex

\section{Motivating applications}\label{sec:motivating_applications}
In this section, we consider three concrete applications in federated learning \cite{Kairouz_survery2020,Li_survery2020} and multi-task reinforcement learning \cite{Zeng_MTRL_2020}, which can be formulated as problem \eqref{prob:main_eq}. Thus, these problems can be  
solved by Algorithm \ref{alg:Dist_non_SA}, and therefore, one can use our results to provide its theoretical guarantees.

\subsection{Federated learning}\label{sec:motivating_applications:FL}
In federated learning, multiple agents (clients or workers) collaboratively solve a machine learning problem under the coordination of a centralized server \cite{Kairouz_survery2020,Li_survery2020}. In stead of sharing the data to the server, the agents run local updates of their models (parameters) based on their data, whose results are aggregated at the server with the goal to find the global learning objective. Such an approach has gained much interests recently due to its efficiency in data processing, system privacy, and operating costs. 

A central problem in federated learning is distributed (or federated) optimization problems, where the goal is to solve 
\begin{align}
\underset{\theta\in\Rset^{d}}{\text{minimize }}G(\theta) \triangleq \frac{1}{N}\sum_{i=1}^{N}G_{i}(\theta).\label{sec:application:FL:obj}    
\end{align}
Here each $G_{i}(\theta) = \Eset_{X_{i}\sim \pi_{i}}[G_{i}(\theta,X_{i})]$ is the loss function and $\pi_{i}$ is the distribution of the data located at agent $i$. For $i\neq j$, $\pi_{i}$ and $\pi_{j}$ are very different, which is referred to as data heterogeneity across the agents. The most popular method for solving \eqref{sec:application:FL:obj} is the so-called local stochastic gradient descent ({\sf SGD}) \cite{Doan_random_projection_2018,stich2018local,McMahan_LocalSGD17,Woodworth_LocalSGD2020,Haddadpour_LocalSGD2019,Karimireddy_SCAFFOLDSC2019,Khaled_LocalSGD2019}, which can be viewed as a variant of Algorithm \ref{alg:Dist_non_SA}. In particular, let $F_{i}(\theta) = \nabla G_{i}(\theta)$ and $F_{i}(\theta,X_{i})$ is its stochastic (sub)gradient $\nabla G_{i}(\theta,X_{i})$. Then at each iteration $k\geq0$, each agent $i$ initialize $\theta_{i}^{k,0} = \btheta^{k}$ and runs $H$ steps of local {\sf SGD} to update $\theta_{i}^{k,t}$. These values are then aggregated by the server to update for $\btheta^{k+1}$, i.e.,
\begin{align*}
    \begin{aligned}
&\text{Agent i: }\quad \theta_{i}^{k,t+1} = \theta_{i}^{k,t}  - \alpha_{k+t}\nabla G_{i}(\theta_{i}^{k,t},X_{i}^{k+t}),\qquad \theta_{i}^{k,0} = \btheta^{k},\; t \in[0,H-1]\\
&\text{Server: }\qquad \btheta^{k+1} = \frac{1}{N}\sum_{i=1}^{N}\theta_{i}^{k,H}.
    \end{aligned}
\end{align*}
In federated optimization literature, it is often assumed that the sequence of samples $\{X_{i}^{k}\}$ are sampled i.i.d from $\pi_{i}$ and the resulting stochastic gradients are unbiased, i.e., $\nabla G_{i}(\theta) = \Eset[\nabla G_{i}(\theta,X_{i}^{k})]$. In addition, the variance of these samples is assumed to be bounded. These assumptions are obviously the special case of the ones considered in this paper.         

\subsection{Distributed multi-task reinforcement learning}\label{sec:motivating_applications:MTRL}
We consider a multi-task reinforcement learning problem over a network of $N$ agents operating in $N$ different environments modeled by Markov random processes ({MDPs}). Here, each environment represents a task assigned to each agent. We assume that the agents can communicate directly with a centralized coordinator. This is a distributed multi-task reinforcement learning problem ({\sf MTRL}) over multi-agent systems, which can be mathematically characterized by using $N$ different MDPs as follows.

Let $\Mcal_{i} = (\Scal_{i},\Acal_{i},\Pcal_{i},\Rcal_{i},\gamma_{i})$ be a $5$-tuple representing the discounted reward {\sf MDP} at agent $i$, where $\Scal_{i},\Acal_{i},$ and $\Pcal_{i}$ are the set of states, action, and transition probability matrices, respectively. In addition, $\Rcal_{i}$ is the reward function and $\gamma_{i}\in(0,1)$ is the discount factor. Note that the set of states and actions at the agents can (partially) overlap with each other, and we denote them by $\Scal = \cup_{i}\Scal_{i}$ and $\Acal = \cup_{i}\Acal_{i}$. We focus on randomized stationary policies ({\sf RSPs}), where a policy $\pi$ assigns to each $s\in\Scal$ a probability distribution $\pi(\cdot|s)$ over $\Acal$.   

Given a policy $\pi$, let $V^{\pi}_i$ be the value function associated with the $i$-th environment,
\begin{align*}
V_i^{\pi}(s_{i}) = \Eset\left[\sum_{k=0}^{\infty}\gamma_{i}^{k}\Rcal_{i}(s^{k}_{i},a_{i}^{k})\,|\,s_{i}^{0} = s_{i}\right],\qquad a_{i}^{k} \sim \pi(\cdot|s_{i}^{k}).  
\end{align*}
Similarly, we denote by $Q_i^{\pi}$ the $Q$-function in the $i$-th environment
\begin{align*}
Q_i^{\pi}(s_{i},a_{i}) = \Eset\left[\sum_{k=0}^{\infty}\gamma_{i}^{k}\Rcal(s_{i}^{k},a_{i}^{k})\,|\,s_{i}^{0} = s_{i}, a_{i}^{0} = a_{i}\right].
\end{align*} 
Let $\rho_{i}$ be an initial state distribution over $\Scal_i$ and with some abuse of notation we denote the long-term reward associated with this distribution as $V_{i}^{\pi}(\rho_{i}) = \Eset_{s_{i}\sim\rho_{i}}\left[V_{i}^{\pi}(s_{i})\right]$. 
The goal of the agents is to cooperatively find a policy $\pi^*$ that maximizes the total accumulative discounted rewards 
\begin{align}
\max_{\pi} V(\pi;\boldsymbol{\rho}) \triangleq \sum_{i=1}^{N}V_{i}^{\pi}(\rho_{i}),
\quad \boldsymbol{\rho} = \left[\begin{array}{c} \rho_1 \\ \vdots \\ \rho_N\end{array}\right].   
\label{sec:motivating_applications:MTRL:obj}
\end{align}
Treating each of the environments as independent {\sf RL} problems would produce different policies $\pi_{i}^*$, each maximizing their respective $V_i^\pi$.  The goal of {\sf MTRL} is to to find a single $\pi^*$ that balances the performance across all environments. In the following, we present two fundamental problems in this area, which can be formulated as problem \eqref{prob:Fi}. As a consequence, we can apply Algorithm \ref{alg:Dist_non_SA} to solve these problems in a distributed manner. 

\subsubsection{Distributed {\sf TD}$(\lambda)$ with linear function approximation}
One of the most fundamental problems in {\sf RL} is the so-called policy evaluation problems, where the goal is to estimate the value function $V^{\pi}$ associated with a stationary policy $\pi$. This problem arises as a subproblem in {\sf RL} policy search methods, including policy
iteration and actor-critic methods. Our focus here is to study the multi-task variant of the policy evaluation problems, that is, we want to estimate the sum of the value functions $V_{i}^{\pi}$ of a stationary policy $\pi$. In addition, we study this problem when the number of state space $\Scal$ is very large, motivating us to use function approximation. We consider the linear function approximation $\tilde{V}_{i}^{\theta}$ of $V_{i}^{\pi}$ parameterized by a weight vector $\theta\in\Rset^{L}$ and given as
\begin{align*}
\tilde{V}_{i}^{\theta}(s) = \sum_{\ell=1}^{L}\theta_{\ell}\phi_{i,\ell}(s),\qquad \forall s\in\Scal,
\end{align*}  
for a given set of $L$ basis vectors $\phi_{i,\ell}:\Scal\rightarrow\Rset$, $\ell\in\{1,\ldots,L\}$, where some examples of how to choose these vectors can be found in \cite{KonidarisOT2011_Linear_Func_approx}. Here we assume that $\phi_{i,\ell}(s) = 0$ if $s\notin \Scal_{i}$, implying $\tilde{V}_{i}^{\theta}(s) = 0$. We are interested in the case  $L\ll M = |\Scal|$. Our goal is to find $\theta^*$ such that it provides a good approximation of the sum of the value functions at the agents, i.e., 
\begin{align*}
\sum_{i=1}^{N}V_{i}^{\pi} \approx \sum_{i=1}^{N}\Phi_{i}\theta^*,
\end{align*}
where $\Phi_{i}\in\Rset^{M\times L}$ is the feature matrix, whose $i$-th row is $\phi_{i}(s)\in\Rset^{L}$, the feature vector of the agent $i$
\begin{align*}
\phi_{i}(s) = (\phi_{i,1}(s),\ldots,\phi_{i,L}(s))^{T}\in\Rset^{L}.
\end{align*}

\textbf{Distributed {\sf TD}$(\lambda)$}. For solving this problem, we consider a distributed variant of {\sf TD}$(\lambda)$, originally studied in \cite{Sutton1988_TD} and analyzed explicitly in \cite{Tsitsiklis1997_TD, Bhandari2018_FiniteTD,SrikantY2019_FiniteTD}. For simplicity, we consider the case $\lambda = 0$, while the case of $\lambda\in[0,1]$ can be done in a similar manner.  In particular, each agent $i$ maintains an estimate $\theta_{i}$ of $\theta^*$ and the centralized coordinator maintains $\btheta$, the averages of these $\theta_{i}$. At each iteration $k\geq 0$, each agent $i$ initialize $\theta_{i}^{k,0} = \btheta^{k}$ and runs $H$ steps of local {\sf TD}$(0)$ to update $\theta_{i}^{k,t}$. These values are then aggregated by the server to update for $\btheta^{k+1}$, i.e., set $\theta_{i}^{k,0} = \btheta^{k}$ and for all $t \in[0,H-1]$
\begin{align}
    \begin{aligned}
&\text{Agent i: }\quad \theta_{i}^{k,t+1} = \theta_{i}^{k,t}  + \alpha_{k+t}\left(\Rcal_{i}^{k+t} + \gamma \phi_{i}(s_{i}^{k+t+1})^T\theta_{i}^{k,t} - \phi_{i}(s_{i}^{k+t})^T\theta_{i}^{k,t}\right)\phi_{i}(s_{i}^{k+t})^T\\
&\text{Server: }\qquad \btheta^{k+1} = \frac{1}{N}\sum_{i=1}^{N}\theta_{i}^{k,H},
    \end{aligned}\label{sec:motivating_applications:MTRL:DTD}
\end{align}
where $\Rcal_{i}^{k} = \Rcal_{i}(s_{i}^{k},a_{i}^{k})$ and $\{s_{i}^{k},s_{i}^{k+1},\Rcal_{i}^{k}\}$ is the data tuple observed at time $k$ at agent $i$. 

Let $\{X_{i}^{k}\} = \left\{(s_{i}^{k},s_{i}^{k+1},a_{i}^{k})\right\}$ be a Markov chain. The update above can be viewed as a local stochastic approximation for finding the root of some linear operator. Indeed, let $\Abf_{i}(X_{i}^{k})$ and $b_{i}(X_{i}^{k})$ be defined as 
\begin{align*}
    \begin{aligned}
        \Abf_{i}(X_{i}^{k}) &= \phi(s_{i}^{k})(\gamma\phi(s_{i}^{k+1}) - \phi(s_{i}^{k})^T,\\
        b_{i}(X_{i}^{k}) &= R_{i}^{k}\phi(s_{i}^{k}).
    \end{aligned}
\end{align*}
Moreover, let $\pi_{i}$ be the stationary distribution of the underlying Markov chain $\{X_{i}^{k}\}$ and 
\begin{align*}
\Abf_{i} = \Eset_{\pi_{i}}\left[\Abf_{i}(X_{i}^{k})\right],\qquad b_{i} = \Eset_{\pi_{i}}\left[b_{i}(X_{i}^{k})\right].
\end{align*}
Thus, in this case if we consider 
\begin{align*}
F_{i}(\theta_{i}^{k};X_{i}^{k}) = - \Abf_{i}(X_{i}^{k})\theta_{i}^{k}  - b_{i}(X_{i}^{k}),    
\end{align*}
then \eqref{sec:motivating_applications:MTRL:DTD} is a variant of Algorithm \ref{alg:Dist_non_SA} where each $F_{i}$ is linear in $\theta$. In this case, the local {\sf TD}$(0)$ seeks to find $\theta^*$ satisfying
\begin{align*} \sum_{i=1}^{N}\Abf_{i}\theta^* + b_{i} = 0.
\end{align*}
To establish the convergence of \eqref{sec:motivating_applications:MTRL:DTD} the following conditions are assumed in the literature \cite{Tsitsiklis1997_TD,Bhandari2018_FiniteTD,SrikantY2019_FiniteTD}. 
\begin{assump}\label{sec:motivating_applications:MTRL:assump:reward}
The instantaneous rewards at the agents are uniformly bounded, i.e., there exist a constant $R$ such that $|\,\Rcal_{i}(s,s')\,|\leq R$, for all $s,s\in\Scal$ and $i\in[N]$. 
\end{assump}
\begin{assump}\label{sec:motivating_applications:MTRL:assump:features}
For each $i\in[N]$, the feature vectors $\{\phi_{i,\ell}\}$, for all $\ell\in\{1,\ldots,L\}$, are linearly independent, i.e., the matrix $\Phi_{i}$ has full column rank. In addition, we assume that all feature vectors $\phi_{i}(s)$ are uniformly bounded, i.e., $\|\phi_{i}(s)\|\leq 1$.  
\end{assump}
\begin{assump}\label{sec:motivating_applications:MTRL:assump:Markov}
Each Markov chain $\{X_{i}^{k}\}$ is irreducible and aperiodic.
\end{assump}
Under Assumption \ref{sec:motivating_applications:MTRL:assump:reward}--\ref{sec:motivating_applications:MTRL:assump:Markov} one can verify that Assumptions \ref{assump:ergodicity}--\ref{assump:strong_monotone} hold \cite{Tsitsiklis1997_TD}. For example, under these assumptions each $\Abf_{i}$ is a negative definite matrix, i.e., $x^T\Abf_{i}x < 0$ for all $x$.

\subsubsection{Distributed Q-learning with linear function approximation}
In this section, we consider a distributed variant of the classic Q-learning method \cite{Watkins_Qlearning} for solving problem \eqref{sec:motivating_applications:MTRL:obj}. Similar to the case of {\sf TD}$(\lambda)$, we focus on the linear function approximation $\tilde{Q}_{i}^{\theta}$ of $Q_{i}$ parameterized by a weight vector $\theta\in\Rset^{L}$ and defined as
\begin{align*}
\tilde{Q}_{i}^{\theta}(s,a) = \sum_{\ell=1}^{L}\theta_{\ell}\phi_{i,\ell}(s,a),\qquad \forall (s,a)\in\Scal\times\Acal
\end{align*}  
for a given set of $L$ basis vectors $\phi_{i,\ell}:\Scal\times\Acal\rightarrow\Rset$, $\ell\in{1,\ldots,L}$. We assume again that $\phi_{i,\ell}(s,a) = 0$ is either $s\notin\Scal_{i}$ or $a\notin\Acal_{i}$, implying $\tilde{Q}_{i}^{\theta}(s,a) = 0$. Let $\Phi_{i}\in\Rset^{|\Scal||\Acal|\times L}$ be the feature matrix, whose $i$-th row is $\phi_{i}(s,a) = (\phi_{i,1}(s,a),\ldots,\phi_{i,L}(s,a)^T$. The the goal of distributed Q-learning is to solve 
\begin{align*}
\sum_{i=1}^{N}\Phi_{i}\theta = \sum_{i=1}^{N}\Pi \Tcal_{i}[\Phi_{i}\theta], 
\end{align*}
where $\Pi$ denotes the projection on the linear subspace of the feature vectors and $\Tcal_{i}$ is the Bellman operator associated with $Q$ function at environment $i$; see for example \cite{ChenZDMC2019}. 

For solving this problem, we consider a distributed variant of Q-learning \cite{ChenZDMC2019,melo_Qlearning_2008,Le_Qlearning_2019}. In particular, each agent $i$ maintains an estimate $\theta_{i}$ of $\theta^*$ and the centralized coordinator maintains $\btheta$, the averages of these $\theta_{i}$. At each iteration $k\geq 0$, each agent $i$ initialize $\theta_{i}^{k,0} = \btheta^{k}$ and runs $H$ steps of local Q-learning to update $\theta_{i}^{k,t}$. These values are then aggregated by the server to update for $\btheta^{k+1}$, i.e., set $\theta_{i}^{k,0} = \btheta^{k}$ and for all $t \in[0,H-1]$
\begin{align}
&\text{Agent i: }\quad \theta_{i}^{k,t+1} = \theta_{i}^{k,t}  + \alpha_{k+t}\left(\Rcal_{i}^{k+t} + \gamma \max_{a'}\phi_{i}(s_{i}^{k+t+1},a')^T\theta_{i}^{k,t} - \phi_{i}(s_{i}^{k+t},a^{k+t})^T\theta_{i}^{k,t}\right)\phi_{i}(s_{i}^{k+t},a^{k+t})^T\notag\\
&\text{Server: }\qquad \btheta^{k+1} = \frac{1}{N}\sum_{i=1}^{N}\theta_{i}^{k,H},\label{sec:motivating_applications:MTRL:DQ}
\end{align}
where $\Rcal_{i}^{k} = \Rcal_{i}(s_{i}^{k},a_{i}^{k})$ and $\{s_{i}^{k},s_{i}^{k+1},a_{i}^{k}\}$ is the sample trajectory generated by some behavior policy $\sigma_{i}$ at agent $i$. Let $X_{i}^{k} = \{s_{i}^{k},a_{i}^{k},s_{i}^{k+1}\}$ be a Markov chain. We denote the nonlinear mapping $F_{i}$ as 
\begin{align*}
F_{i}(\theta;X^{k}) = \phi_{i}(s^{k},a^{k})\left[\Rcal(s^{k},a^{k}) + \gamma\max_{a}\phi_{i}(s^{k+1},a)^T\theta - \phi_{i}(s^{k},a^{k})^T\theta\right],    
\end{align*}
and $F_{i}(\theta) = \Eset_{\pi_{i}}\left[F_{i}(\theta;X^{k})\right]$, where $\pi_{i}$ is the stationary distribution of $\{X_{i}^{k}\}$. Then the goal of \eqref{sec:motivating_applications:MTRL:DQ} is to find $\theta^*$ such that 
\begin{align*}
    \sum_{i=1}^{N} F_{i}(\theta^*) = 0. 
 \end{align*}
Under proper assumptions, for example see \cite[Theorem 1]{ChenZDMC2019}, one can verify that Assumptions \ref{assump:ergodicity}--\ref{assump:strong_monotone} hold. Thus, we can apply our results in Theorems \ref{thm:const} and \ref{thm:tv} to derive the finite-time performance bound for distributed Q-learning in \eqref{sec:motivating_applications:MTRL:DQ}.

%% file: appendix.tex

\appendix

\section{Proofs of Lemmas \ref{sec:analysis:const:lem:theta_i_bound}--\ref{sec:analysis:tv:lem:bias}}

\subsection{Proof of Lemma \ref{sec:analysis:const:lem:theta_i_bound}}
We first show \eqref{sec:analysis:const:lem:theta_i_bound:ineq1}. Indeed, by \eqref{alg:theta_i} and \eqref{lem:bounded_Fi:ineq} we have for any $t\in[0,H-1]$  
\begin{align}
\|\theta_{i}^{k,t+1} - \theta_{i}^{k,t}\| = \alpha \|F_{i}(\theta_{i}^{k,t};X_{i}^{k+t})\| \leq B\alpha (\|\theta_{i}^{k,t}\|+1),\label{sec:analysis:const:lem:theta_i_bound:eq1a}    
\end{align}
which gives 
\begin{align*}
\|\theta_{i}^{k,t+1}\| &\leq (1+B\alpha)\|\theta_{i}^{k,t}\| + B\alpha \leq (1+B\alpha)^{t+1}\|\theta_{i}^{k,0}\| + B\alpha \sum_{u = 0}^{t}(1+B\alpha)^{t-u}.
\end{align*}
Using the relation $1+x\leq e^{x}$ for all $x\geq 0$ into the preceding equation gives \eqref{sec:analysis:const:lem:theta_i_bound:ineq1}, i.e., 
\begin{align*}
\|\theta_{i}^{k,t+1}\| 
&\leq e^{B\alpha(t+1)}\|\theta_{i}^{k,0}\| + B\alpha te^{B\alpha t}\leq e^{B\alpha H}\|\theta_{i}^{k,0}\| +  BH\alpha   e^{B\alpha H} \leq 2\|\btheta^{k}\| + 2BH\alpha \leq 2\|\btheta^{k}\| + 1,
\end{align*}
where the second inequality is due to \eqref{sec:analysis:const:stepsize}, i.e., $HB\alpha \leq \log(2)/2\tau(\alpha)\leq \log(2)$, and recall that $\theta_{i}^{k,0} = \btheta^{k}$. Next, using \eqref{sec:analysis:const:lem:theta_i_bound:eq1a} and \eqref{sec:analysis:const:lem:theta_i_bound:ineq1} we obtain for all $t\in[0,H-1]$ 
\begin{align}
\|\theta_{i}^{k,t+1}-\theta_{i}^{k,t}\| \leq 2B\alpha \|\btheta^{k}\| + 2B^2H\alpha^2 + B\alpha,    \label{lem:btheta_k_tau:eq1c}
\end{align}
which implies \eqref{sec:analysis:const:lem:theta_i_bound:ineq2}, i.e., for all $t\in[1,H]$ we have
\begin{align*}
\|\theta_{i}^{k,t} - \btheta^{k}\| &= \left\|\sum_{u=0}^{t-1}\theta_{i}^{k,u+1} - \theta_{i}^{k,u}\right\|\leq \sum_{u = 0}^{t-1}\|\theta_{i}^{k,u+1} - \theta_{i}^{k,u}\|\notag\\
&\leq 2B\alpha t \|\btheta^{k}\| + 2B^2H\alpha^2 t + B\alpha t\leq 2BH\alpha\|\btheta^{k}\| + 2B^2H^2\alpha^2 + BH\alpha\notag\\
&\leq 2BH\alpha\|\btheta^{k}\|  + 2BH\alpha,
\end{align*}
where the last inequality is due to \eqref{sec:analysis:const:stepsize}. 

\subsection{Proof of Lemma \ref{sec:analysis:const:lem:btheta_k_tau}}
We first show \eqref{sec:analysis:const:lem:btheta_k_tau:ineq1}. Using \eqref{sec:analysis:const:btheta_k} and \eqref{lem:bounded_Fi:ineq} we have 
\begin{align*}
\|\btheta^{k+1}\| - \|\btheta^{k}\| &\leq \|\btheta^{k+1} - \btheta^{k}\| \leq \frac{\alpha}{N}\sum_{i=1}^N\sum_{t=0}^{H-1}\|F_{i}(\theta_{i}^{k,t};X_{i}^{k+t})\|\notag\\ 
&\leq    \frac{\alpha}{N}\sum_{i=1}^N\sum_{t=0}^{H-1} B\left(\|\theta_{i}^{k,t}\| + 1\right) \leq BH\alpha\left(\|\btheta^{k}\| + 1\right) + \frac{\alpha}{N}\sum_{i=1}^N\sum_{t=0}^{H-1} B\|\theta_{i}^{k,t}-\btheta^{k}\| ,
\end{align*}
which when using \eqref{sec:analysis:const:lem:theta_i_bound:ineq2} and $BH\alpha\leq \log(2)/2$ (from \eqref{sec:analysis:const:stepsize})  gives
\begin{align}
\|\btheta^{k+1}\| - \|\btheta^{k}\| &\leq \|\btheta^{k+1}-\btheta^{k}\| \leq BH\alpha\left(\|\btheta^{k}\|+1\right) + BH\alpha\left(2BH\alpha\|\btheta^{k}\| + 2BH\alpha\right)\notag\\
&\leq 2BH\alpha\|\btheta^{k}\| + 2BH\alpha,\label{sec:analysis:const:lem:btheta_k_tau:eq1d}
\end{align}
The preceding relation yields
\begin{align*}
\|\btheta^{k+1}\| \leq (1+2BH\alpha)\|\btheta^{k}\| + 2BH\alpha.
\end{align*}
Using the relation $1+x \leq e^{x}$ for all $x\geq 0$, the equation above gives for all $t\in[k-\tau(\alpha),k-1]$
\begin{align*}
\|\btheta^{t}\| &\leq (1 + 2BH\alpha)^{t-k + \tau(\alpha)}\|\btheta^{k-\tau(\alpha)}\| + 2BH\sum_{u = k-\tau(\alpha)}^{t-1}\alpha(1 + 2BH\alpha)^{t-u-1}\notag\\
&\leq (1 + 2BH\alpha)^{\tau(\alpha)}\|\btheta^{k-\tau(\alpha)}\| + 2BH\alpha\tau(\alpha)(1 + 2BH\alpha)^{t-1-k+\tau(\alpha)}\notag\\
&\leq e^{2BH\alpha\tau(\alpha)}\|\btheta^{k-\tau(\alpha)}\| + 2BH\alpha\tau(\alpha)(1+2BH\alpha)^{\tau(\alpha)} \leq e^{2BH\alpha\tau(\alpha)}\|\btheta^{k-\tau(\alpha)}\| + 2BH\alpha\tau(\alpha)e^{2BH\alpha\tau(\alpha)}\notag\\
&\leq 2\|\btheta^{k-\tau(\alpha)}\| + 4BH\alpha\tau(\alpha),
\end{align*}
where the last inequality is due to \eqref{sec:analysis:const:stepsize}, i.e.,
$2HB\tau(\alpha)\alpha\leq \log(2).$ Using the preceding relation we have from \eqref{sec:analysis:const:lem:btheta_k_tau:eq1d} 
\begin{align*}
\|\btheta^{k}-\btheta^{k-\tau(\alpha)}\|&\leq \sum_{t=k-\tau(\alpha)}^{k-1}\|\btheta^{t+1} - \btheta^{t}\|\leq \sum_{t=k-\tau(\alpha)}^{k-1}2BH\alpha(\|\btheta^{t}\| + 1)\notag\\
&\leq 2BH\alpha\sum_{t=k-\tau(\alpha)}^{k-1}\left(2\|\btheta^{k-\tau(\alpha)}\| + 4BH\alpha\tau(\alpha)\right) + 2BH\alpha\tau(\alpha)\notag\\
&\leq 4BH\alpha\tau(\alpha)\|\btheta^{k-\tau(\alpha)}\| + 4BH\alpha\tau(\alpha),
\end{align*}
where the last inequality is due to \eqref{sec:analysis:const:stepsize}, i.e., $2HB\tau(\alpha)\alpha\leq \log(2)\leq 1/2$. Using the preceding inequality and the triangle inequality yields
\begin{align*}
\|\btheta^{k}-\btheta^{k-\tau(\alpha_{k})}\|&\leq 4BH\alpha\tau(\alpha)\|\btheta^{k} - \btheta^{k-\tau(\alpha)}\| + 4BH\alpha\tau(\alpha)\|\btheta^{k}\| + 4BH\alpha\tau(\alpha)\notag\\
&\leq \frac{2}{3}\|\btheta^{k} - \btheta^{k-\tau(\alpha)}\| + 4BH\alpha\tau(\alpha)\|\btheta^{k}\| + 4BH\alpha\tau(\alpha),
\end{align*}
where the last inequality we use \eqref{sec:analysis:const:stepsize} to  have $2BH\tau(\alpha_{k})\alpha\leq \log(2)\leq 1/3$. Rearranging the equation above yields \eqref{sec:analysis:const:lem:btheta_k_tau:ineq1}
\begin{align*}
 \|\btheta^{k}-\btheta^{k-\tau(\alpha)}\| \leq  12BH\alpha\tau(\alpha)\|\btheta^{k}\| + 12BH\alpha\tau(\alpha)\leq 2\|\btheta^{k}\| + 2.  
\end{align*}
Taking square on both sides of the preceding relation and using the Cauchy-Schwarz inequality yield \eqref{sec:analysis:const:lem:btheta_k_tau:ineq2}
\begin{align*}
\|\btheta^{k}-\btheta^{k-\tau(\alpha)}\|^2 \leq    288B^2H^2\alpha^2\tau^2(\alpha)\|\btheta^{k}\|^2 + 288B^2H^2\alpha^2\tau^2(\alpha)\leq 8\|\btheta^{k}\|^2 + 8. 
\end{align*}

\subsection{Proof of Lemma \ref{sec:analysis:const:lem:bias}}
Consider 
\begin{align}
&- \sum_{i=1}^{N}\sum_{t=0}^{H-1}\left\langle\btheta^{k}-\theta^*,F_{i}(\theta_{i}^{k,t};X_{i}^{k+t})-F_{i}(\btheta^{k})\right\rangle\notag\\ 
&= - \sum_{i=1}^{N}\sum_{t=0}^{H-1}\left\langle\btheta^{k}-\btheta^{k-\tau(\alpha)},F_{i}(\theta_{i}^{k,t};X_{i}^{k+t})-F_{i}(\btheta^{k})\right\rangle - \sum_{i=1}^{N}\sum_{t=0}^{H-1} \left\langle\btheta^{k-\tau(\alpha)}-\theta^*,F_{i}(\theta_{i}^{k,t};X_{i}^{k+t})-F_{i}(\btheta^{k})\right\rangle\notag\\
&= - \sum_{i=1}^{N}\sum_{t=0}^{H-1}\left\langle\btheta^{k}-\btheta^{k-\tau(\alpha)},F_{i}(\theta_{i}^{k,t};X_{i}^{k+t})-F_{i}(\btheta^{k})\right\rangle\notag\\ 
&\quad - \sum_{i=1}^{N}\sum_{t=0}^{H-1} \left\langle\btheta^{k-\tau(\alpha)}-\theta^*,F_{i}(\theta_{i}^{k-\tau(\alpha),t};X_{i}^{k+t})-F_{i}(\theta_{i}^{k-\tau(\alpha),t})\right\rangle\notag\\
&\quad -  \sum_{i=1}^{N}\sum_{t=0}^{H-1}\left\langle\btheta^{k-\tau(\alpha)}-\theta^*,F_{i}(\theta_{i}^{k,t};X_{i}^{k+t})-F_{i}(\theta_{i}^{k-\tau(\alpha),t};X_{i}^{k+t})\right\rangle\notag\\
&\quad -  \sum_{i=1}^{N}\sum_{t=0}^{H-1}\left\langle\btheta^{k-\tau(\alpha)}-\theta^*,F_{i}(\theta_{i}^{k-\tau(\alpha),t})-F_{i}(\btheta^{k-\tau(\alpha)})\right\rangle\notag\\
&\quad -  \sum_{i=1}^{N}\sum_{t=0}^{H-1}\left\langle\btheta^{k-\tau(\alpha)}-\theta^*,F_{i}(\btheta^{k-\tau(\alpha)}) - F_{i}(\btheta^{k})\right\rangle.\label{sec:analysis:const:lem:bias:Eq1}
\end{align}
We first consider the second term on the right-hand side of \eqref{sec:analysis:const:lem:bias:Eq1}. Let $\Fcal_{k}$ be the set containing all the information generated by Algorithm \ref{alg:Dist_non_SA} up to time $k$. Then, using \eqref{lem:mixing_time:ineq} we have
\begin{align*}
& -\sum_{i=1}^{N}\sum_{t=0}^{H-1}\Eset\left[  \left\langle\btheta^{k-\tau(\alpha)}-\theta^*,F_{i}(\theta_{i}^{k-\tau(\alpha),t};X_{i}^{k+t})-F_{i}(\theta_{i}^{k-\tau(\alpha),t})\right\rangle\,|\, \Fcal_{k+t-\tau(\alpha)}\right]\notag\\
&= -\sum_{i=1}^{N}\sum_{t=0}^{H-1}\left\langle\btheta^{k-\tau(\alpha)}-\theta^*,\Eset\left[ F_{i}(\theta_{i}^{k-\tau(\alpha),t};X_{i}^{k+t})-F_{i}(\theta_{i}^{k-\tau(\alpha),t})\,|\, \Fcal_{k+t-\tau(\alpha)}\right]\right\rangle\notag\\
&\leq \sum_{i=1}^{N}\sum_{t=0}^{H-1}\left\|\btheta^{k-\tau(\alpha)}-\theta^*\right\|\left|\Eset\left[ F_{i}(\theta_{i}^{k-\tau(\alpha),t};X_{i}^{k+t})-F_{i}(\theta_{i}^{k-\tau(\alpha),t})\,|\, \Fcal_{k+t-\tau(\alpha)}\right]\right|\notag\\
&\leq \sum_{i=1}^{N}\sum_{t=0}^{H-1}\alpha\left\|\btheta^{k-\tau(\alpha)}-\theta^*\right\|\left(\left\|\theta_{i}^{k-\tau(\alpha),t}\right\|+1\right) = NH\alpha\left\|\btheta^{k-\tau(\alpha)}-\theta^*\right\|\left(\left\|\theta_{i}^{k-\tau(\alpha),t}\right\|+1\right),
\end{align*}
which by using the triangle inequality and \eqref{sec:analysis:const:lem:theta_i_bound:ineq1} yields
\begin{align}
& -\sum_{i=1}^{N}\sum_{t=0}^{H-1}\Eset\left[  \left\langle\btheta^{k-\tau(\alpha)}-\theta^*,F_{i}(\theta_{i}^{k-\tau(\alpha),t};X_{i}^{k+t})-F_{i}(\theta_{i}^{k-\tau(\alpha),t})\right\rangle\,|\, \Fcal_{k+t-\tau(\alpha)}\right]\notag\\
&\leq NH\alpha\left(\|\btheta^{k} - \btheta^{k-\tau(\alpha)}\| + \|\btheta^{k} - \theta^*\|\right)\left(2\|\btheta^{k-\tau(\alpha)}\| + 2BH\alpha   +1\right)\notag\\
&\leq NH\alpha \left(\|\btheta^{k} - \btheta^{k-\tau(\alpha)}\| + \|\btheta^{k} - \theta^*\|\right)\left(2\|\btheta^{k} - \btheta^{k-\tau(\alpha)}\| + 2\|\btheta^{k}\| + 2\right)\notag\\
&\leq 2NH\alpha\left(2\|\btheta{k}\| + 2 + \|\btheta^{k} - \theta^*\|\right)\left(3\|\btheta^{k}\| + 3\right)\notag\\
&\leq 6NH\alpha\left(3\|\btheta^{k}-\theta^*\| + 2 + 2\|\theta^*\|\right)\left(\|\btheta^{k}-\theta^*\| + 1 + \|\theta^*\|\right)\notag\\
&\leq 18NH\alpha\left(\|\btheta^{k}-\theta^*\| + 1 + \|\theta^*\|\right)^2\leq 36NH\alpha\left(\|\btheta^{k}-\theta^*\|\right)^2 + 36NH\alpha \left(1 + \|\theta^*\|\right)^2,\label{sec:analysis:const:lem:bias:Eq1a}
\end{align}
where the third inequality is due to \eqref{sec:analysis:const:lem:btheta_k_tau:ineq1} and the last inequality is due to the Cauchy-Schwarz inequality. Next, we consider the third term on the right-hand side of \eqref{sec:analysis:const:lem:bias:Eq1}. Indeed, using \eqref{assump:Lipschitz_sample:ineq} we have
\begin{align}
&-  \sum_{i=1}^{N}\sum_{t=0}^{H-1}\left\langle\btheta^{k-\tau(\alpha)}-\theta^*,F_{i}(\theta_{i}^{k,t};X_{i}^{k+t})-F_{i}(\theta_{i}^{k-\tau(\alpha),t};X_{i}^{k+t})\right\rangle\notag\\ 
&\leq L\sum_{i=1}^{N}\sum_{t=0}^{H-1}\|\btheta^{k-\tau(\alpha)}-\theta^*\|\|\theta_{i}^{k,t} - \theta_{i}^{k-\tau(\alpha),t}\|\notag\\
&\leq L\sum_{i=1}^{N}\sum_{t=0}^{H-1}\|\btheta^{k-\tau(\alpha)}-\theta^*\|\left(\|\theta_{i}^{k,t} - \btheta^{k}\| + \|\btheta^{k} - \btheta^{k-\tau(\alpha)}\| + \|\btheta^{k-\tau(\alpha)} - \theta_{i}^{k-\tau(\alpha),t}\|\right).\label{sec:analysis:const:lem:bias:Eq1b}
\end{align}
Similarly, using \eqref{assump:Lipschitz_sample:ineq} we consider the last two terms on the right-hand sides of \eqref{sec:analysis:const:lem:bias:Eq1} 
\begin{align*}
 & - \sum_{i=1}^{N}\sum_{t=0}^{H-1}\left( \left\langle\btheta^{k-\tau(\alpha)}-\theta^*,F_{i}(\theta_{i}^{k-\tau(\alpha),t})-F_{i}(\btheta^{k-\tau(\alpha)})\right\rangle +  \left\langle\btheta^{k-\tau(\alpha)}-\theta^*,F_{i}(\btheta^{k-\tau(\alpha)}) - F_{i}(\btheta^{k})\right\rangle\right)\notag\\
&\leq L\sum_{i=1}^{N}\sum_{t=0}^{H-1}\|\btheta^{k-\tau(\alpha)}-\theta^*\|\left(\|\theta_{i}^{k-\tau(\alpha),t} - \btheta^{k-\tau(\alpha)}\| + \|\btheta^{k} - \btheta^{k-\tau(\alpha)}\|\right), 
\end{align*}
which by adding to \eqref{sec:analysis:const:lem:bias:Eq1b} yields
\begin{align*}
 & - \sum_{i=1}^{N}\sum_{t=0}^{H-1}\left( \left\langle\btheta^{k-\tau(\alpha)}-\theta^*,F_{i}(\theta_{i}^{k-\tau(\alpha),t})-F_{i}(\btheta^{k-\tau(\alpha)})\right\rangle +  \left\langle\btheta^{k-\tau(\alpha)}-\theta^*,F_{i}(\btheta^{k-\tau(\alpha)}) - F_{i}(\btheta^{k})\right\rangle\right)\notag\\
 &\qquad -  \sum_{i=1}^{N}\sum_{t=0}^{H-1}\left\langle\btheta^{k-\tau(\alpha)}-\theta^*,F_{i}(\theta_{i}^{k,t};X_{i}^{k+t})-F_{i}(\theta_{i}^{k-\tau(\alpha),t};X_{i}^{k+t})\right\rangle\notag\\ 
&\leq L\sum_{i=1}^{N}\sum_{t=0}^{H-1}\|\btheta^{k-\tau(\alpha)}-\theta^*\|\left(\|\theta_{i}^{k,t} - \btheta^{k}\| + 2\|\theta_{i}^{k-\tau(\alpha),t} - \btheta^{k-\tau(\alpha)}\| + 2\|\btheta^{k} - \btheta^{k-\tau(\alpha)}\|\right).
\end{align*}
Using \eqref{sec:analysis:const:lem:theta_i_bound:ineq2}, \eqref{sec:analysis:const:lem:btheta_k_tau:ineq2}, and the triangle inequality into the preceding relation yields
\begin{align}
& - \sum_{i=1}^{N}\sum_{t=0}^{H-1}\left( \left\langle\btheta^{k-\tau(\alpha)}-\theta^*,F_{i}(\theta_{i}^{k-\tau(\alpha),t})-F_{i}(\btheta^{k-\tau(\alpha)})\right\rangle +  \left\langle\btheta^{k-\tau(\alpha)}-\theta^*,F_{i}(\btheta^{k-\tau(\alpha)}) - F_{i}(\btheta^{k})\right\rangle\right)\notag\\
 &\qquad -  \sum_{i=1}^{N}\sum_{t=0}^{H-1}\left\langle\btheta^{k-\tau(\alpha)}-\theta^*,F_{i}(\theta_{i}^{k,t};X_{i}^{k+t})-F_{i}(\theta_{i}^{k-\tau(\alpha),t};X_{i}^{k+t})\right\rangle\notag\\  
 &\leq L\sum_{i=1}^{N}\sum_{t=0}^{H-1}\left(\|\btheta^{k-\tau(\alpha)}-\btheta^{k}\| + \|\btheta^{k}-\theta^*\|\right)\notag\\
 &\hspace{2.5cm}\times\left(2BH\alpha\|\btheta^{k}\| + 4BH\alpha\|\btheta^{k-\tau(\alpha)}\| + 6BH\alpha + 24BH\alpha\tau(\alpha)\|\btheta^{k}\| + 18BH\alpha\tau(\alpha)\right)\notag\\
&\leq LNH\left(2\|\btheta^{k}\| + 2 + \|\btheta^{k} - \theta^*\| \right)\left(26BH\alpha\tau(\alpha)\|\btheta^{k}\| + 24BH\alpha\tau(\alpha) + 4BH\alpha\|\btheta^{k-\tau(\alpha)}-\btheta^{k}\| + 4BH\alpha\|\btheta^{k}\|  \right)\notag\\
&\leq LNH\left(3\|\btheta^{k}-\theta^*\| + 2 + 2\|\theta^*\| \right)\left(38BH\alpha\tau(\alpha)\|\btheta^{k}\| + 32BH\alpha\tau(\alpha)\right)\notag\\
&\leq LNH\left(3\|\btheta^{k}-\theta^*\| + 2 + 2\|\theta^*\| \right)\left(38BH\alpha\tau(\alpha)\|\btheta^{k}-\theta^*\| + 38BH(\|\theta^*\|+1)\alpha\tau(\alpha)\right)\notag\\
&\leq 114LBNH^2\alpha\tau(\alpha)\left(\|\btheta^{k}-\theta^*\| + \|\theta^*\|+1\right)^2\notag\\
&\leq 228LBNH^2\alpha\tau(\alpha)\left(\|\btheta^{k}-\theta^*\|\right)^2 + 228LBNH^2\alpha\tau(\alpha)\left(\|\theta^*\|+1\right)^2.  \label{sec:analysis:const:lem:bias:Eq1c}   
\end{align}
Finally, we consider the first term on the right-hand side of \eqref{sec:analysis:const:lem:bias:Eq1}. Using \eqref{sec:analysis:const:lem:theta_i_bound:ineq1}, \eqref{sec:analysis:const:lem:btheta_k_tau:ineq1}, and \eqref{lem:bounded_Fi:ineq} we consider 
\begin{align}
&- \sum_{i=1}^{N}\sum_{t=0}^{H-1}\left\langle\btheta^{k}-\btheta^{k-\tau(\alpha)},F_{i}(\theta_{i}^{k,t};X_{i}^{k+t})-F_{i}(\btheta^{k})\right\rangle \leq B\sum_{i=1}^{N}\sum_{t=0}^{H-1}\left\|\btheta^{k}-\btheta^{k-\tau(\alpha)}\right\|\left(\|\theta_{i}^{k,t}\|  + \|\btheta^{k}\| + 2 \right)\notag\\
&\leq B\sum_{i=1}^{N}\sum_{t=0}^{H-1}\left(12BH\alpha\tau(\alpha)\|\btheta^{k}\| + 12BH\alpha\tau(\alpha)\right)\left(3\|\btheta^{k}\| + 2 + 2BH\alpha\right)\notag\\
&\leq 12NB^2H^2\alpha\tau(\alpha)\left(\|\btheta^{k}\| + 1\right)(3\|\btheta^{k}\| + 3)\leq 36NB^2H^2\alpha\tau(\alpha)(\|\btheta^{k}\|+1)^2\notag\\
&\leq 36NB^2H^2\alpha\tau(\alpha)(\|\btheta^{k}-\theta^*\| + \|\theta^*\| + 1)^2\notag\\
&\leq 72NB^2H^2\alpha\tau(\alpha)\|\btheta^{k}-\theta^*\|^2 + 72NB^2H^2(1+\|\theta^*\|)^2\alpha\tau(\alpha),
\label{sec:analysis:const:lem:bias:Eq1d}      
\end{align}
where in the third inequality we use \eqref{sec:analysis:const:stepsize} to have $2BH\alpha \leq 1$. Finally, taking the expectation on both sides of \eqref{sec:analysis:const:lem:bias:Eq1} and using \eqref{sec:analysis:const:lem:bias:Eq1a}, \eqref{sec:analysis:const:lem:bias:Eq1c}, and \eqref{sec:analysis:const:lem:bias:Eq1d} yields \eqref{sec:analysis:const:lem:bias:ineq}
\begin{align*}
& - \sum_{i=1}^{N}\sum_{t=0}^{H-1}\Eset\left[\left\langle\btheta^{k}-\theta^*,F_{i}(\theta_{i}^{k,t};X_{i}^{k+t})-F_{i}(\btheta^{k})\right\rangle\right]\notag\\
&\leq 36NH\alpha\Eset\left[\|\btheta^{k}-\theta^*\|^2\right] + 36NH\alpha \left(1 + \|\theta^*\|\right)^2 + 228LBNH^2\alpha\tau(\alpha)\Eset\left[\|\btheta^{k}-\theta^*\|^2\right]  \notag\\ 
&\qquad + 228LBNH^2\alpha\tau(\alpha)\left(\|\theta^*\|+1\right)^2 + 72NB^2H^2\alpha\tau(\alpha)\Eset\left[\|\btheta^{k}-\theta^*\|^2\right] + 72NB^2H^2(1+\|\theta^*\|)^2\alpha\tau(\alpha) \notag\\
&\leq 36NH\alpha\Eset\left[\|\btheta^{k}-\theta^*\|^2\right] + 36NH\alpha \left(1 + \|\theta^*\|\right)^2\notag\\ 
&\qquad + 12(19L+6B)NBH^2\alpha\tau(\alpha)\Eset\left[\|\btheta^{k}-\theta^*\|^2\right] + 12(19L+6B)NBH^2\left(1 + \|\theta^*\|\right)^2\alpha\tau(\alpha).
\end{align*}

\subsection{Proof of Lemma \ref{sec:analysis:tv:lem:theta_i_bound}}
We first show \eqref{sec:analysis:tv:lem:theta_i_bound:ineq1}. Indeed, by \eqref{alg:theta_i} and \eqref{lem:bounded_Fi:ineq} we have for any $t\in[0,H-1]$  
\begin{align}
\|\theta_{i}^{k,t+1} - \theta_{i}^{k,t}\| = \alpha_{k+t} \|F_{i}(\theta_{i}^{k,t};X_{i}^{k+t})\| \leq B\alpha_{k+t} (\|\theta_{i}^{k,t}\|+1),\label{sec:analysis:tv:lem:theta_i_bound:eq1a}    
\end{align}
which gives 
\begin{align*}
\|\theta_{i}^{k,t+1}\| &\leq (1+B\alpha_{k+t})\|\theta_{i}^{k,t}\| + B\alpha_{k+t} \leq \prod_{u=0}^{t}(1+B\alpha_{k+u})\|\theta_{i}^{k,0}\| + B \sum_{u = 0}^{t}\alpha_{k+u}\prod_{\ell = u+1}^{t}(1+B\alpha_{k+\ell}).
\end{align*}
Using the relation $1+x\leq e^{x}$ for all $x\geq 0$ into the preceding equation and since $\alpha_{k}$ is decreasing we obtain \eqref{sec:analysis:tv:lem:theta_i_bound:ineq1}, i.e., for $t\in[0,H-1]$
\begin{align*}
\|\theta_{i}^{k,t+1}\| 
&\leq \exp\left\{B\sum_{u=0}^{t}\alpha_{k+u}\right\}\|\theta_{i}^{k,0}\| + B \sum_{u = 0}^{t}\alpha_{k+u}\exp\left\{B\sum_{\ell = u+1}^{t}\alpha_{k+\ell}\right\}\notag\\
&\leq \exp\{BH\alpha_{k}\}\|\theta_{i}^{k,0}\| + B\sum_{u=0}^{t}\alpha_{k+u}\exp\{BH\alpha_{k}\}\notag\\
&\leq 2\|\btheta^{k}\| + 2BH\alpha_{k} ,
\end{align*}
where the second inequality is  \eqref{sec:analysis:tv:stepsize}, i.e., $HB\alpha_{k} \leq \log(2)$, and recall that $\theta_{i}^{k,0} = \btheta^{k}$. Next, using \eqref{sec:analysis:tv:lem:theta_i_bound:eq1a} and \eqref{sec:analysis:tv:lem:theta_i_bound:ineq1} and since $\alpha_{k}$ is decreasing we obtain for all $t\in[0,H-1]$ 
\begin{align}
\|\theta_{i}^{k,t+1}-\theta_{i}^{k,t}\| \leq B\alpha_{k}\left(\|\theta_{i}^{k,t}\|+1\right) \leq 2B\alpha_{k} \|\btheta^{k}\| + 2B^2H\alpha_{k}^2 + B\alpha_{k},    \label{lem:btheta_k_tau:eq1c}
\end{align}
which implies \eqref{sec:analysis:const:lem:theta_i_bound:ineq2}, i.e., for all $t\in[1,H]$ we have
\begin{align*}
\|\theta_{i}^{k,t} - \btheta^{k}\| &= \left\|\sum_{u=0}^{t-1}\theta_{i}^{k,u+1} - \theta_{i}^{k,u}\right\|\leq \sum_{u = 0}^{t-1}\|\theta_{i}^{k,u+1} - \theta_{i}^{k,u}\|\notag\\
&\leq 2B\alpha_{k} t \|\btheta^{k}\| + 2B^2H\alpha_{k}^2 t + B\alpha_{k} t\leq 2BH\alpha_{k}\|\btheta^{k}\| + 2B^2H^2\alpha_{k}^2 + BH\alpha_{k}\notag\\
&\leq 2BH\alpha_{k}\|\btheta^{k}\|  + 2BH\alpha_{k},
\end{align*}
where the last inequality is due to \eqref{sec:analysis:tv:stepsize}. 

\subsection{Proof of Lemma \ref{sec:analysis:tv:lem:btheta_k_tau}}
We first show \eqref{sec:analysis:tv:lem:btheta_k_tau:ineq1}. Using \eqref{sec:analysis:tv:btheta_k} and \eqref{lem:bounded_Fi:ineq} we have 
\begin{align*}
\|\btheta^{k+1}\| - \|\btheta^{k}\| &\leq \|\btheta^{k+1} - \btheta^{k}\| \leq \frac{1}{N}\sum_{i=1}^N\sum_{t=0}^{H-1}\alpha_{k+t}\|F_{i}(\theta_{i}^{k,t};X_{i}^{k+t})\|\notag\\ 
&\leq    \frac{1}{N}\sum_{i=1}^N\sum_{t=0}^{H-1} B\alpha_{k+t}\left(\|\theta_{i}^{k,t}\| + 1\right) \leq BH\alpha_{k}\left(\|\btheta^{k}\| + 1\right) + \frac{1}{N}\sum_{i=1}^N\sum_{t=0}^{H-1} B\alpha_{k+t}\|\theta_{i}^{k,t}-\btheta^{k}\| ,
\end{align*}
which when using \eqref{sec:analysis:tv:lem:theta_i_bound:ineq2} and $BH\alpha_{k}\leq \log(2)/2$ (from \eqref{sec:analysis:tv:stepsize})  gives
\begin{align}
\|\btheta^{k+1}\| - \|\btheta^{k}\| &\leq \|\btheta^{k+1}-\btheta^{k}\| \leq BH\alpha_{k}\left(\|\btheta^{k}\|+1\right) + BH\alpha_{k}\left(2BH\alpha_{k}\|\btheta^{k}\| + 2BH\alpha_{k}\right)\notag\\
&\leq 2BH\alpha_{k}\|\btheta^{k}\| + 2BH\alpha_{k},\label{sec:analysis:tv:lem:btheta_k_tau:eq1d}
\end{align}
The preceding relation yields
\begin{align*}
\|\btheta^{k+1}\| \leq (1+2BH\alpha_{k})\|\btheta^{k}\| + 2BH\alpha_{k}.
\end{align*}
Using the relation $1+x \leq e^{x}$ for all $x\geq 0$, the equation above gives for all $t\in[k-\tau(\alpha_{k}),k-1]$
\begin{align*}
\|\btheta^{t+1}\| &\leq \prod_{u=k-\tau(\alpha_{k})}^{t}(1 + 2BH\alpha_{u})\|\btheta^{k-\tau(\alpha_{k})}\| + 2BH\sum_{u = k-\tau(\alpha_{k})}^{t-1}\alpha_{u}\prod_{\ell=u+1}^{t}(1 + 2BH\alpha_{\ell})\notag\\
&\leq \exp\left\{\sum_{u = k -\tau(\alpha_{k})}^{t}2BH\alpha_{u}\right\}\|\btheta^{k-\tau(\alpha_{k})}\| + 2BH\sum_{u=k-\tau(\alpha_{k})}^{t-1}\alpha_{u}\exp\left\{\sum_{\ell = u+1}^{t}2BH\alpha_{\ell}\right\}\notag\\
&\leq 2\|\btheta^{k-\tau(\alpha_{k})}\| + 4BH\sum_{u=k-\tau(\alpha_{k})}^{t-1}\alpha_{u},
\end{align*}
where the last inequality is due to \eqref{sec:analysis:tv:stepsize}, i.e.,
$2HB\alpha_{k;\tau(\alpha_{k})}\leq \log(2).$ Using the preceding relation we have from \eqref{sec:analysis:tv:lem:btheta_k_tau:eq1d} 
\begin{align*}
\|\btheta^{k}-\btheta^{k-\tau(\alpha_{k})}\|&\leq \sum_{t=k-\tau(\alpha_{k})}^{k-1}\|\btheta^{t+1} - \btheta^{t}\|\leq \sum_{t=k-\tau(\alpha_{k})}^{k-1}2BH\alpha_{t}(\|\btheta^{t}\| + 1)\notag\\
&\leq 2BH\sum_{t=k-\tau(\alpha)}^{k-1}\alpha_{t}\left(2\|\btheta^{k-\tau(\alpha)}\| + 4BH\sum_{u=k-\tau(\alpha_{k})}^{t-1}\alpha_{u}\right) + 2BH\alpha_{k;\tau(\alpha_{k})}\notag\\
&\leq 4BH\alpha_{k;\tau(\alpha_{k})}\|\btheta^{k-\tau(\alpha)}\| + 4BH\alpha_{k;\tau(\alpha_{k})},
\end{align*}
where the last inequality is due to \eqref{sec:analysis:tv:stepsize}, i.e., $2HB\alpha_{k;\tau(\alpha_{k})}\leq \log(2)\leq 1/2$. Using the preceding inequality and the triangle inequality yields
\begin{align*}
\|\btheta^{k}-\btheta^{k-\tau(\alpha_{k})}\|&\leq 4BH\alpha_{k;\tau(\alpha_{k})}\|\btheta^{k} - \btheta^{k-\tau(\alpha)}\| + 4BH\alpha_{k;\tau(\alpha_{k})}\|\btheta^{k}\| + 4BH\alpha_{k;\tau(\alpha_{k})}\notag\\
&\leq \frac{2}{3}\|\btheta^{k} - \btheta^{k-\tau(\alpha)}\| + 4BH\alpha_{k;\tau(\alpha_{k})}\|\btheta^{k}\| + 4BH\alpha_{k;\tau(\alpha_{k})},
\end{align*}
where the last inequality we use \eqref{sec:analysis:tv:stepsize} to  have $2BH\alpha_{k;\tau(\alpha_{k})}\leq \log(2)\leq 1/3$. Rearranging the equation above yields \eqref{sec:analysis:const:lem:btheta_k_tau:ineq1}
\begin{align*}
 \|\btheta^{k}-\btheta^{k-\tau(\alpha)}\| \leq  12BH\alpha_{k;\tau(\alpha_{k})}\|\btheta^{k}\| + 12BH\alpha_{k;\tau(\alpha_{k})}\leq 2\|\btheta^{k}\| + 2.  
\end{align*}
Taking square on both sides of the preceding relation and using the Cauchy-Schwarz inequality yield \eqref{sec:analysis:const:lem:btheta_k_tau:ineq2}
\begin{align*}
\|\btheta^{k}-\btheta^{k-\tau(\alpha)}\|^2 \leq    288B^2H^2\alpha_{k;\tau(\alpha_{k})}^2\|\btheta^{k}\|^2 + 288B^2H^2\alpha_{k;\tau(\alpha_{k})}^2\leq 8\|\btheta^{k}\|^2 + 8. 
\end{align*}

\subsection{Proof of Lemma \ref{sec:analysis:tv:lem:bias}}
Consider 
\begin{align}
&- \sum_{i=1}^{N}\sum_{t=0}^{H-1}\alpha_{k+t}\left\langle\btheta^{k}-\theta^*,F_{i}(\theta_{i}^{k,t};X_{i}^{k+t})-F_{i}(\btheta^{k})\right\rangle\notag\\ 
&= - \sum_{i=1}^{N}\sum_{t=0}^{H-1}\alpha_{k+t}\left\langle\btheta^{k}-\btheta^{k-\tau(\alpha_{k})},F_{i}(\theta_{i}^{k,t};X_{i}^{k+t})-F_{i}(\btheta^{k})\right\rangle\notag\\ 
&\qquad  - \sum_{i=1}^{N}\sum_{t=0}^{H-1} \alpha_{k+t}\left\langle\btheta^{k-\tau(\alpha_{k})}-\theta^*,F_{i}(\theta_{i}^{k,t};X_{i}^{k+t})-F_{i}(\btheta^{k})\right\rangle\notag\\
&= - \sum_{i=1}^{N}\sum_{t=0}^{H-1}\alpha_{k+t}\left\langle\btheta^{k}-\btheta^{k-\tau(\alpha_{k})},F_{i}(\theta_{i}^{k,t};X_{i}^{k+t})-F_{i}(\btheta^{k})\right\rangle\notag\\ 
&\quad - \sum_{i=1}^{N}\sum_{t=0}^{H-1} \alpha_{k+t}\left\langle\btheta^{k-\tau(\alpha_{k})}-\theta^*,F_{i}(\theta_{i}^{k-\tau(\alpha_{k}),t};X_{i}^{k+t})-F_{i}(\theta_{i}^{k-\tau(\alpha_{k}),t})\right\rangle\notag\\
&\quad -  \sum_{i=1}^{N}\sum_{t=0}^{H-1}\alpha_{k+t}\left\langle\btheta^{k-\tau(\alpha_{k})}-\theta^*,F_{i}(\theta_{i}^{k,t};X_{i}^{k+t})-F_{i}(\theta_{i}^{k-\tau(\alpha_{k}),t};X_{i}^{k+t})\right\rangle\notag\\
&\quad -  \sum_{i=1}^{N}\sum_{t=0}^{H-1}\alpha_{k+t}\left\langle\btheta^{k-\tau(\alpha_{k})}-\theta^*,F_{i}(\theta_{i}^{k-\tau(\alpha_{k}),t})-F_{i}(\btheta^{k-\tau(\alpha_{k})})\right\rangle\notag\\
&\quad -  \sum_{i=1}^{N}\sum_{t=0}^{H-1}\alpha_{k+t}\left\langle\btheta^{k-\tau(\alpha_{k})}-\theta^*,F_{i}(\btheta^{k-\tau(\alpha_{k})}) - F_{i}(\btheta^{k})\right\rangle.\label{sec:analysis:tv:lem:bias:Eq1}
\end{align}
We first consider the second term on the right-hand side of \eqref{sec:analysis:tv:lem:bias:Eq1}. Let $\Fcal_{k}$ be the set containing all the information generated by Algorithm \ref{alg:Dist_non_SA} up to time $k$. Then, using \eqref{lem:mixing_time:ineq} we have
\begin{align*}
& -\sum_{i=1}^{N}\sum_{t=0}^{H-1}\alpha_{k+t}\Eset\left[  \left\langle\btheta^{k-\tau(\alpha_{k})}-\theta^*,F_{i}(\theta_{i}^{k-\tau(\alpha_{k}),t};X_{i}^{k+t})-F_{i}(\theta_{i}^{k-\tau(\alpha_{k}),t})\right\rangle\,|\, \Fcal_{k+t-\tau(\alpha_{k})}\right]\notag\\
&= -\sum_{i=1}^{N}\sum_{t=0}^{H-1}\alpha_{k+t}\left\langle\btheta^{k-\tau(\alpha_{k})}-\theta^*,\Eset\left[ F_{i}(\theta_{i}^{k-\tau(\alpha_{k}),t};X_{i}^{k+t})-F_{i}(\theta_{i}^{k-\tau(\alpha_{k}),t})\,|\, \Fcal_{k+t-\tau(\alpha_{k})}\right]\right\rangle\notag\\
&\leq \sum_{i=1}^{N}\sum_{t=0}^{H-1}\alpha_{k+t}\left\|\btheta^{k-\tau(\alpha_{k})}-\theta^*\right\|\left|\Eset\left[ F_{i}(\theta_{i}^{k-\tau(\alpha_{k}),t};X_{i}^{k+t})-F_{i}(\theta_{i}^{k-\tau(\alpha_{k}),t})\,|\, \Fcal_{k+t-\tau(\alpha_{k})}\right]\right|\notag\\
&\leq \sum_{i=1}^{N}\sum_{t=0}^{H-1}\alpha_{k+t}\alpha_{k}\left\|\btheta^{k-\tau(\alpha_{k})}-\theta^*\right\|\left(\left\|\theta_{i}^{k-\tau(\alpha_{k}),t}\right\|+1\right) = NH\alpha_{k}^2\left\|\btheta^{k-\tau(\alpha_{k})}-\theta^*\right\|\left(\left\|\theta_{i}^{k-\tau(\alpha_{k}),t}\right\|+1\right),
\end{align*}
which by using the triangle inequality and \eqref{sec:analysis:tv:lem:theta_i_bound:ineq1} yields
\begin{align}
& -\sum_{i=1}^{N}\sum_{t=0}^{H-1}\alpha_{k+t}\Eset\left[  \left\langle\btheta^{k-\tau(\alpha_{k})}-\theta^*,F_{i}(\theta_{i}^{k-\tau(\alpha_{k}),t};X_{i}^{k+t})-F_{i}(\theta_{i}^{k-\tau(\alpha_{k}),t})\right\rangle\,|\, \Fcal_{k+t-\tau(\alpha_{k})}\right]\notag\\
&\leq NH\alpha_{k}^2\left(\|\btheta^{k} - \btheta^{k-\tau(\alpha_{k})}\| + \|\btheta^{k} - \theta^*\|\right)\left(2\|\btheta^{k-\tau(\alpha_{k})}\| + 2BH\alpha_{k}   +1\right)\notag\\
&\leq NH\alpha_{k}^2 \left(\|\btheta^{k} - \btheta^{k-\tau(\alpha_{k})}\| + \|\btheta^{k} - \theta^*\|\right)\left(2\|\btheta^{k} - \btheta^{k-\tau(\alpha_{k})}\| + 2\|\btheta^{k}\| + 2\right)\notag\\
&\leq 2NH\alpha_{k}^2\left(2\|\btheta{k}\| + 2 + \|\btheta^{k} - \theta^*\|\right)\left(3\|\btheta^{k}\| + 3\right)\notag\\
&\leq 6NH\alpha_{k}^2\left(3\|\btheta^{k}-\theta^*\| + 2 + 2\|\theta^*\|\right)\left(\|\btheta^{k}-\theta^*\| + 1 + \|\theta^*\|\right)\notag\\
&\leq 18NH\alpha_{k}^2\left(\|\btheta^{k}-\theta^*\| + 1 + \|\theta^*\|\right)^2\leq 36NH\alpha_{k}^2\left(\|\btheta^{k}-\theta^*\|\right)^2 + 36NH\alpha_{k}^2 \left(1 + \|\theta^*\|\right)^2,\label{sec:analysis:tv:lem:bias:Eq1a}
\end{align}
where the third inequality is due to \eqref{sec:analysis:tv:lem:btheta_k_tau:ineq1} and the last inequality is due to the Cauchy-Schwarz inequality. Next, we consider the third term on the right-hand side of \eqref{sec:analysis:tv:lem:bias:Eq1}. Indeed, using \eqref{assump:Lipschitz_sample:ineq} we have
\begin{align}
&-  \sum_{i=1}^{N}\sum_{t=0}^{H-1}\alpha_{k+t}\left\langle\btheta^{k-\tau(\alpha_{k})}-\theta^*,F_{i}(\theta_{i}^{k,t};X_{i}^{k+t})-F_{i}(\theta_{i}^{k-\tau(\alpha_{k}),t};X_{i}^{k+t})\right\rangle\notag\\ 
&\leq L\sum_{i=1}^{N}\sum_{t=0}^{H-1}\alpha_{k+t}\|\btheta^{k-\tau(\alpha_{k})}-\theta^*\|\|\theta_{i}^{k,t} - \theta_{i}^{k-\tau(\alpha_{k}),t}\|\notag\\
&\leq L\sum_{i=1}^{N}\sum_{t=0}^{H-1}\alpha_{k+t}\|\btheta^{k-\tau(\alpha_{k})}-\theta^*\|\left(\|\theta_{i}^{k,t} - \btheta^{k}\| + \|\btheta^{k} - \btheta^{k-\tau(\alpha_{k})}\| + \|\btheta^{k-\tau(\alpha_{k})} - \theta_{i}^{k-\tau(\alpha_{k}),t}\|\right).\label{sec:analysis:tv:lem:bias:Eq1b}
\end{align}
Similarly, using \eqref{assump:Lipschitz_sample:ineq} we consider the last two terms on the right-hand sides of \eqref{sec:analysis:tv:lem:bias:Eq1} 
\begin{align*}
 & - \sum_{i=1}^{N}\sum_{t=0}^{H-1}\alpha_{k+t}\left( \left\langle\btheta^{k-\tau(\alpha_{k})}-\theta^*,F_{i}(\theta_{i}^{k-\tau(\alpha_{k}),t})-F_{i}(\btheta^{k-\tau(\alpha_{k})})\right\rangle +  \left\langle\btheta^{k-\tau(\alpha_{k})}-\theta^*,F_{i}(\btheta^{k-\tau(\alpha_{k})}) - F_{i}(\btheta^{k})\right\rangle\right)\notag\\
&\leq L\sum_{i=1}^{N}\sum_{t=0}^{H-1}\alpha_{k+t}\|\btheta^{k-\tau(\alpha_{k})}-\theta^*\|\left(\|\theta_{i}^{k-\tau(\alpha_{k}),t} - \btheta^{k-\tau(\alpha_{k})}\| + \|\btheta^{k} - \btheta^{k-\tau(\alpha_{k})}\|\right), 
\end{align*}
which by adding to \eqref{sec:analysis:tv:lem:bias:Eq1b} yields
\begin{align*}
 & - \sum_{i=1}^{N}\sum_{t=0}^{H-1}\alpha_{k+t}\left( \left\langle\btheta^{k-\tau(\alpha_{k})}-\theta^*,F_{i}(\theta_{i}^{k-\tau(\alpha_{k}),t})-F_{i}(\btheta^{k-\tau(\alpha_{k})})\right\rangle +  \left\langle\btheta^{k-\tau(\alpha_{k})}-\theta^*,F_{i}(\btheta^{k-\tau(\alpha_{k})}) - F_{i}(\btheta^{k})\right\rangle\right)\notag\\
 &\qquad -  \sum_{i=1}^{N}\sum_{t=0}^{H-1}\alpha_{k+t}\left\langle\btheta^{k-\tau(\alpha_{k})}-\theta^*,F_{i}(\theta_{i}^{k,t};X_{i}^{k+t})-F_{i}(\theta_{i}^{k-\tau(\alpha_{k}),t};X_{i}^{k+t})\right\rangle\notag\\ 
&\leq L\sum_{i=1}^{N}\sum_{t=0}^{H-1}\alpha_{k+t}\|\btheta^{k-\tau(\alpha_{k})}-\theta^*\|\left(\|\theta_{i}^{k,t} - \btheta^{k}\| + 2\|\theta_{i}^{k-\tau(\alpha_{k}),t} - \btheta^{k-\tau(\alpha_{k})}\| + 2\|\btheta^{k} - \btheta^{k-\tau(\alpha_{k})}\|\right).
\end{align*}
Using \eqref{sec:analysis:tv:lem:theta_i_bound:ineq2}, \eqref{sec:analysis:tv:lem:btheta_k_tau:ineq1}, and the triangle inequality into the preceding relation yields
\begin{align}
& - \sum_{i=1}^{N}\sum_{t=0}^{H-1}\alpha_{k+t}\left( \left\langle\btheta^{k-\tau(\alpha_{k})}-\theta^*,F_{i}(\theta_{i}^{k-\tau(\alpha_{k}),t})-F_{i}(\btheta^{k-\tau(\alpha_{k})})\right\rangle +  \left\langle\btheta^{k-\tau(\alpha_{k})}-\theta^*,F_{i}(\btheta^{k-\tau(\alpha_{k})}) - F_{i}(\btheta^{k})\right\rangle\right)\notag\\
 &\qquad -  \sum_{i=1}^{N}\sum_{t=0}^{H-1}\alpha_{k+t}\left\langle\btheta^{k-\tau(\alpha_{k})}-\theta^*,F_{i}(\theta_{i}^{k,t};X_{i}^{k+t})-F_{i}(\theta_{i}^{k-\tau(\alpha_{k}),t};X_{i}^{k+t})\right\rangle\notag\\  
 &\leq L\sum_{i=1}^{N}\sum_{t=0}^{H-1}\alpha_{k+t}\left(\|\btheta^{k-\tau(\alpha_{k})}-\btheta^{k}\| + \|\btheta^{k}-\theta^*\|\right)\left(2BH\alpha_{k}\|\btheta^{k}\| +2BH\alpha_{k} + 4BH\alpha_{k}\|\btheta^{k-\tau(\alpha_{k})}\| + 4BH\alpha_{k-\tau(\alpha_{k})}\right) +\notag\\
 &\qquad + L\sum_{i=1}^{N}\sum_{t=0}^{H-1}\alpha_{k+t}\left(\|\btheta^{k-\tau(\alpha_{k})}-\btheta^{k}\| + \|\btheta^{k}-\theta^*\|\right)\left(24BH\alpha_{k;\tau(\alpha_{k})}\|\btheta^{k}\| + 24BH\alpha_{k;\tau(\alpha_{k})}\right)\notag\\
&\leq LNBH^2\alpha_{k}\left(2\|\btheta^{k}\| + 2 + \|\btheta^{k} - \theta^*\| \right)\left(26\alpha_{k;\tau(\alpha_{k})}\|\btheta^{k}\| + 30\alpha_{k;\tau(\alpha_{k})} + 4\alpha_{k}\|\btheta^{k-\tau(\alpha_{k})}-\btheta^{k}\| + 4\alpha_{k}\|\btheta^{k}\|  \right)\notag\\
&\leq LNBH^2\alpha_{k}\left(3\|\btheta^{k}-\theta^*\| + 2 + 2\|\theta^*\| \right)\left(30\alpha_{k;\tau(\alpha_{k})}\|\btheta^{k}\| + 30\alpha_{k;\tau(\alpha_{k})} + 8\alpha_{k}\|\btheta^{k}\| + 8\alpha_{k} \right)\notag\\
&\leq 38LNBH^2\alpha_{k}\alpha_{k;\tau(\alpha_{k})}\left(3\|\btheta^{k}-\theta^*\| + 2 + 2\|\theta^*\| \right)\left(\|\btheta^{k}\|+1\right)\notag\\
&\leq 114LNBH^2\alpha_{k}\alpha_{k;\tau(\alpha_{k})}\left(\|\btheta^{k}-\theta^*\| + 1 + \|\theta^*\|\right)^2\notag\\
&\leq 228LNBH^2\alpha_{k}\alpha_{k;\tau(\alpha_{k})}\|\btheta^{k}-\theta^*\|^2 + 228LNBH^2\alpha_{k}\alpha_{k;\tau(\alpha_{k})}\left(\|\theta^*\|+1\right)^2.  \label{sec:analysis:tv:lem:bias:Eq1c}   
\end{align}
Finally, we consider the first term on the right-hand side of \eqref{sec:analysis:tv:lem:bias:Eq1}. Using \eqref{sec:analysis:tv:lem:theta_i_bound:ineq1}, \eqref{sec:analysis:tv:lem:btheta_k_tau:ineq1}, and \eqref{lem:bounded_Fi:ineq} we consider 
\begin{align}
&- \sum_{i=1}^{N}\sum_{t=0}^{H-1}\alpha_{k+t}\left\langle\btheta^{k}-\btheta^{k-\tau(\alpha_{k})},F_{i}(\theta_{i}^{k,t};X_{i}^{k+t})-F_{i}(\btheta^{k})\right\rangle \leq B\sum_{i=1}^{N}\sum_{t=0}^{H-1}\alpha_{k+t}\left\|\btheta^{k}-\btheta^{k-\tau(\alpha_{k})}\right\|\left(\|\theta_{i}^{k,t}\|  + \|\btheta^{k}\| + 2 \right)\notag\\
&\leq B\sum_{i=1}^{N}\sum_{t=0}^{H-1}\alpha_{k+t}\left(12BH\alpha_{k;\tau(\alpha_{k})}\|\btheta^{k}\| + 12BH\alpha_{k;\tau(\alpha_{k})}\right)\left(3\|\btheta^{k}\| + 3\right)\notag\\
&\leq 36NB^2H^2\alpha_{k}\alpha_{k;\tau(\alpha_{k})}\left(\|\btheta^{k}\| + 1\right)^2\leq 36NB^2H^2\alpha_{k}\alpha_{k;\tau(\alpha_{k})}(\|\btheta^{k}-\theta^*\| + \|\theta^*\| + 1)^2\notag\\
&\leq 72NB^2H^2\alpha_{k}\alpha_{k;\tau(\alpha_{k})}\|\btheta^{k}-\theta^*\|^2 + 72NB^2H^2(1+\|\theta^*\|)^2\alpha_{k}\alpha_{k;\tau(\alpha_{k})},
\label{sec:analysis:tv:lem:bias:Eq1d}      
\end{align}
where in the third inequality we use \eqref{sec:analysis:tv:stepsize} to have $2BH\alpha \leq 1$. Finally, taking the expectation on both sides of \eqref{sec:analysis:tv:lem:bias:Eq1} and using \eqref{sec:analysis:tv:lem:bias:Eq1a}, \eqref{sec:analysis:tv:lem:bias:Eq1c}, and \eqref{sec:analysis:tv:lem:bias:Eq1d} yields \eqref{sec:analysis:tv:lem:bias:ineq}, i.e.,
\begin{align*}
& - \sum_{i=1}^{N}\sum_{t=0}^{H-1}\alpha_{k+t}\Eset\left[\left\langle\btheta^{k}-\theta^*,F_{i}(\theta_{i}^{k,t};X_{i}^{k+t})-F_{i}(\btheta^{k})\right\rangle\right]\notag\\
&\leq 36NH\alpha_{k}^2\left(\|\btheta^{k}-\theta^*\|\right)^2 + 36NH\alpha_{k}^2 \left(1 + \|\theta^*\|\right)^2 + 228LBNH^2\alpha_{k}\alpha_{k;\tau(\alpha_{k})}\Eset\left[\|\btheta^{k}-\theta^*\|^2\right]  \notag\\ 
&\qquad + 228LBNH^2\alpha_{k}\alpha_{k;\tau(\alpha_{k})}\left(\|\theta^*\|+1\right)^2 + 72NB^2H^2\alpha_{k}\alpha_{k;\tau(\alpha_{k})}\Eset\left[\|\btheta^{k}-\theta^*\|^2\right]\notag\\ 
&\qquad + 72NB^2H^2(1+\|\theta^*\|)^2\alpha_{k}\alpha_{k;\tau(\alpha_{k})} \notag\\
&\leq 36NH\alpha_{k}^2\Eset\left[\|\btheta^{k}-\theta^*\|^2\right] + 36NH\alpha_{k}^2\left(1 + \|\theta^*\|\right)^2\notag\\ 
&\qquad + 12(19L+6B)NBH^2\alpha_{k}\alpha_{k;\tau(\alpha_{k})}\Eset\left[\|\btheta^{k}-\theta^*\|^2\right] + 12(19L+6B)NBH^2\left(1 + \|\theta^*\|\right)^2\alpha_{k}\alpha_{k;\tau(\alpha_{k})}.
\end{align*}